\documentclass{spml_template}

\usepackage[T1]{fontenc}
\usepackage{amsmath}
\usepackage{amsfonts}
\usepackage{amssymb}
\usepackage{array}
\usepackage{booktabs}
\usepackage{colortbl}
\usepackage{longtable}
\usepackage{graphicx}
\usepackage{float}
\usepackage{placeins}
\usepackage{pifont}
\usepackage{algpseudocode}
\floatstyle{ruled}
\newfloat{algorithm}{tbp}{loa}
\floatname{algorithm}{Algorithm}
\usepackage{microtype}
\usepackage{xcolor}
\usepackage{url}
\usepackage{tikz}
\usetikzlibrary{arrows.meta,positioning,backgrounds,fit,calc,shapes.geometric}
\usepackage{pgfplots}
\pgfplotsset{compat=1.18}

\definecolor{reviewfg}{HTML}{CC0000}
\definecolor{mfSlate}{HTML}{5C6470}
\definecolor{mfSlateFill}{HTML}{E7E9EC}
\definecolor{mfPurple}{HTML}{9B5FA0}
\definecolor{mfPurpleFill}{HTML}{F1E5F2}
\definecolor{mfTeal}{HTML}{3E97A8}
\definecolor{mfTealFill}{HTML}{DFF1F4}
\definecolor{mfBlue}{HTML}{3E78C2}
\definecolor{mfBlueFill}{HTML}{E1EBF7}
\definecolor{mfOrange}{HTML}{E0902E}
\definecolor{mfOrangeFill}{HTML}{FAEAD7}
\definecolor{mfGreen}{HTML}{4FA06A}
\definecolor{mfGreenFill}{HTML}{E1F0E6}
\definecolor{mfRose}{HTML}{CC5C5C}
\definecolor{mfRoseFill}{HTML}{F7E2E2}
\definecolor{mfGrid}{HTML}{D7DCE2}

\definecolor{ovInk}{HTML}{303640}
\definecolor{ovTextMuted}{HTML}{59636F}
\definecolor{ovFrame}{HTML}{D6DADF}
\definecolor{ovGroupFill}{HTML}{FAFAF8}
\definecolor{ovInputLine}{HTML}{687C8E}
\definecolor{ovInputFill}{HTML}{E5EDF3}
\definecolor{ovTemplateLine}{HTML}{64866A}
\definecolor{ovTemplateFill}{HTML}{DDEBDD}
\definecolor{ovStereoLine}{HTML}{687EA5}
\definecolor{ovStereoFill}{HTML}{DFE6F4}
\definecolor{ovSpaceLine}{HTML}{AD7645}
\definecolor{ovSpaceFill}{HTML}{F6E1C9}
\definecolor{ovModelLine}{HTML}{846590}
\definecolor{ovModelFill}{HTML}{EADFF0}

\newcommand{\capsupported}{\textcolor{mfGreen}{\ding{51}}}
\newcommand{\capunsupported}{\textcolor{mfRose}{\ding{55}}}
\newcommand{\caprequired}{\textcolor{mfOrange}{\ensuremath{\blacktriangle}}}

\DeclareRobustCommand{\ours}{\textnormal{\textsc{Packora}}}
\DeclareRobustCommand{\model}{\ours}

\title{\model{}: Systematic Design for Generative Molecular Crystal Structure Prediction}

\author[1]{Nayoung Kim}
\author[1]{Kiyoung Seong}
\author[1]{Sungsoo Ahn}

\affiliation[1]{Korea Advanced Institute of Science and Technology (KAIST)}

\metadata[Project Page]{\href{https://nayoung10.github.io/packora/}{\texttt{https://nayoung10.github.io/packora/}}}
\correspondence{\email{nayoungkim@kaist.ac.kr}, \email{sungsoo.ahn@kaist.ac.kr}}

\abstract{
Molecular crystal structure prediction (CSP) is important in pharmaceuticals, agrochemicals, and organic electronics, where subtle differences in molecular conformation and packing can strongly affect material properties. We present \model{}, a flow-based generative model for molecular CSP that jointly predicts atomic coordinates and the lattice from molecular graphs. \model{} supports multi-component and organometallic crystals and can condition on any subset of molecular conformers, stereochemical labels, and space-group information within a single model. Inspired by the CCDC CSP blind test, we evaluate generation and ranking separately, using generation to isolate generator quality and ranking to measure end-to-end performance under a common relaxation and ranking pipeline. We also systematically study architecture, training, conditioning, inference, and scaling, identifying an effective design based on cacheable pairwise reasoning, training objective and numerical solver choices, conditioning dropout, and balanced scaling of pairwise and single representations. \model{} outperforms the baselines on both structure generation and ranking benchmarks, achieving the best matched-budget coverage across all six generation benchmarks, as well as higher experimental-form recovery, lower experimental-form ranks, and faster convergence in ranking.

}

\begin{document}

\maketitle

\begin{figure}[H]
  \centering
  \includegraphics[width=\textwidth]{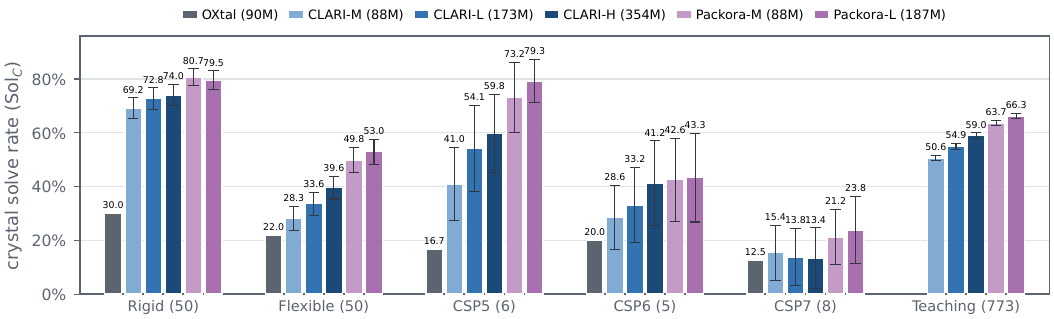}
    \caption{\textbf{Crystal coverage with 30 candidates per target.}
    We report the crystal-level solve rate $\mathrm{Sol}_C$ at a generation budget of
    30 candidates per target. Numbers in parentheses indicate the number of targets in
    each benchmark. For OXtal, we report the published results obtained from 30 generated candidates per target. For CLARI and \model{}, we generate
    1{,}000 candidates per target and report the mean over 5{,}000 bootstrap
    resamples of 30 candidates; error bars show one standard deviation across
    resamples. \model{} achieves the highest solve rate on all six
    benchmarks.}
  \label{fig:benchmark-teaser}
\end{figure}


\section{Introduction}

Molecular crystals are widely used in pharmaceuticals, agrochemicals, and organic electronics~\citep{beran2023frontiers}. Their properties depend strongly on molecular conformation and packing, and different polymorphs of the same chemical system can exhibit distinct solubility, stability, bioavailability, charge transport, and mechanical properties~\citep{price2014predicting,jin2025oxtal,gharakhanyan2025fastcsp}. Molecular crystal structure prediction (CSP) aims to predict possible crystal structures given the molecular components, accelerating drug development and functional materials design.

Molecular CSP generally involves two complementary stages: structure generation and structure ranking~\citep{hunnisett2024seventhranking,hunnisett2024seventhgeneration}. Traditional workflows generate candidate packings using random, quasi-random, or evolutionary search~\citep{li2018genarris,curtis2018gator}, then relax and rank them using computationally expensive energy calculations, often based on density functional theory~\citep{price2014predicting,reilly2016sixth,hunnisett2024seventhranking,hunnisett2024seventhgeneration}. FastCSP primarily accelerates structure ranking by using a machine-learning interatomic potential (MLIP) for structure relaxation and energy calculation, while obtaining initial candidates from a random structure generator~\citep{gharakhanyan2025fastcsp}.

Generative models offer a faster approach to structure generation by learning from experimentally observed crystal structures and concentrating a finite candidate budget on plausible packings. Despite recent progress~\citep{jin2025oxtal,zeng2026molcrystalflow,subramanian2026packflow,lo2026clari}, several gaps remain. Existing methods differ in the molecular information they require and the chemical systems they support, while evaluation protocols vary in candidate budgets and downstream processing, making generation and ranking performance difficult to compare directly. Moreover, the effects of key design choices in architecture, training, conditioning, inference, and scaling remain only partially understood.

In this work, we present \model{}, an all-atom generative model for molecular CSP that jointly predicts atomic coordinates and the unit cell from molecular graphs containing atom types, bond types, and formal charges. \model{} supports multi-component and organometallic crystals and explicitly models hydrogen atoms. A single model can also condition on additional structural information that is often available or readily derivable, such as molecular templates, stereochemical labels, and space-group information, allowing it to adapt to different levels of prior knowledge.

We also introduce a matched two-track evaluation that separates structure generation from structure ranking, following the CCDC CSP blind tests~\citep{hunnisett2024seventhranking,hunnisett2024seventhgeneration}. The generation track measures whether a fixed candidate set contains the experimental structure, without relaxation or energy-based ranking, thereby isolating generator coverage. The ranking track instead processes candidates from each generator through the same relaxation, filtering, deduplication, and lattice-energy ranking pipeline and measures whether the experimental form is ranked near the top. This separation enables direct comparison of generator quality and the downstream utility of generated candidates in practical CSP workflows.

Finally, we systematically study key design choices in architecture, training recipe, conditioning, inference, and scaling. We directly compare DiT with pair-bias attention~\citep{peebles2023scalable}, Pairformer~\citep{Abramson2024}, and Pairmixer~\citep{ouyangzhang2025triangle}, and ablate training choices such as time sampling, translation augmentation, and loss design. These studies show that cacheable pairwise reasoning substantially improves generation quality, careful training and sampling choices matter, and balanced scaling of pairwise and single representations preserves coverage at larger model sizes.

\begin{figure*}[!t]
  \centering
  \includegraphics[width=\textwidth,height=0.82\textheight,keepaspectratio]{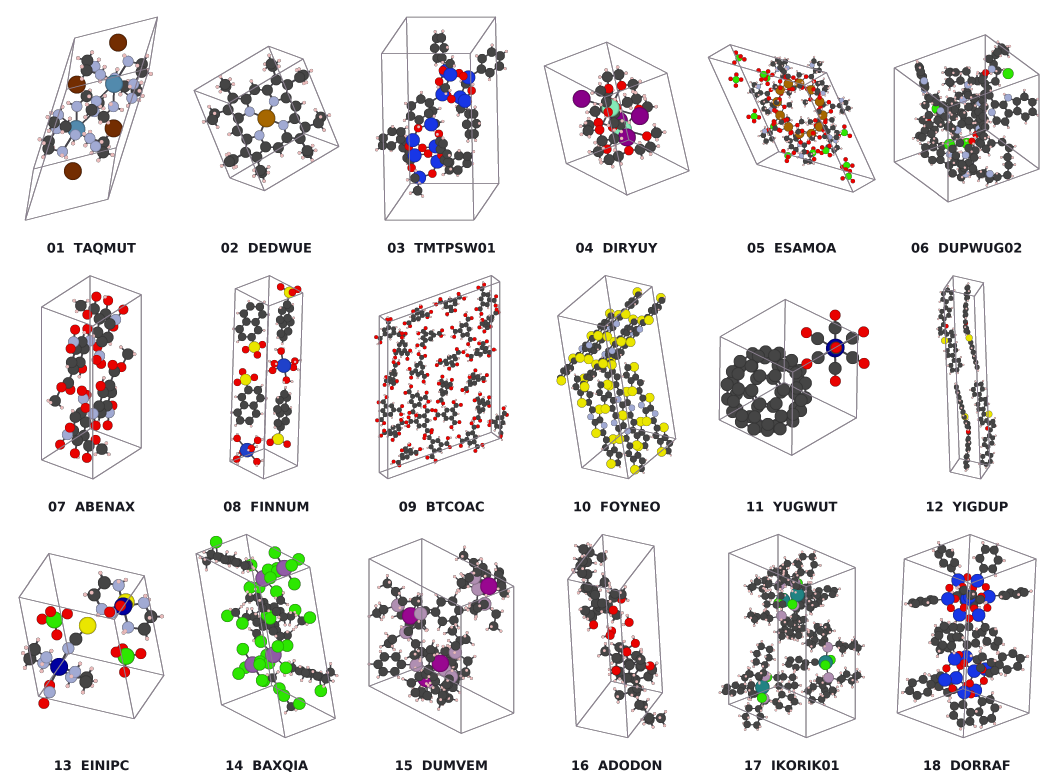}
  \caption{\textbf{Crystal structures predicted by \model{}.}
    Each example is a collision-free prediction from \model{}-L that matches all
    15 molecules of the experimental reference with
    $\mathrm{RMSD}_{15}<2\,\text{\AA}$. These 18 examples illustrate \model{}'s ability to generate diverse crystal structures across molecular sizes, compositions, packing motifs, and unit-cell geometries. Panel labels indicate the corresponding CSD refcodes.}
  \label{fig:strict-structure-examples}
\end{figure*}

For structure generation, across six benchmarks, \model{}-M or \model{}-L achieves the best coverage across candidate budgets and evaluation criteria, with up to a 77.6\% relative improvement over CLARI-H (\Cref{tab:benchmark_solc_main}). When combined with downstream relaxation and ranking, \model{} achieves higher recovery, faster convergence, and lower experimental-form ranks than CLARI-H on both the single-form and polymorph FastCSP benchmarks~\citep{gharakhanyan2025fastcsp}.

In summary, our contributions are:
\begin{itemize}
    \item \textbf{A flexible all-atom generative model for molecular CSP.}
    \model{} jointly generates atomic coordinates and unit cells for diverse molecular systems while supporting optional conditioning on molecular templates, stereochemistry, and space groups.
    \item \textbf{A matched evaluation of generation and ranking.}
    We separately measure finite-budget generator coverage and downstream performance under a common relaxation-and-ranking pipeline.
    \item \textbf{A systematic study of molecular CSP generative modeling.}
    We investigate architecture, training, conditioning, inference, and scaling under a common evaluation protocol and derive an evidence-backed design recipe.
\end{itemize}
\section{Related Work}

\textbf{Search-based molecular CSP.}
Search-based molecular CSP typically involves exploring molecular conformations and crystal packings, followed by geometry relaxation and energy ranking. Structure generation and ranking have long been central components of molecular CSP, with the seventh CCDC blind test explicitly evaluating them in separate phases~\citep{reilly2016sixth,hunnisett2024seventhranking,hunnisett2024seventhgeneration}. Candidate structures are commonly generated across selected space groups by sampling unit-cell parameters and molecular positions and orientations, with flexible molecules requiring additional sampling of intramolecular degrees of freedom~\citep{day2009significant,reilly2016sixth}. Generated structures are then deduplicated or clustered and evaluated hierarchically, often using force fields for initial relaxation and screening followed by dispersion-corrected density functional theory for refinement and ranking~\citep{beran2023frontiers,hunnisett2024seventhranking}. Genarris uses constrained random generation followed by diversity-based down-selection~\citep{li2018genarris}, whereas GAtor uses a first-principles genetic algorithm with molecular-crystal breeding operators and structural niching~\citep{curtis2018gator}. FastCSP accelerates relaxation and ranking stage using a machine-learning interatomic potential, while retaining random structure generation for initial candidates~\citep{gharakhanyan2025fastcsp}.

\textbf{Generative molecular CSP.}
Generative molecular CSP instead learns a distribution over experimental crystal
structures. OXtal is an all-atom diffusion model trained on lattice-free molecular
crops and adapts the AlphaFold3 Pairformer trunk~\citep{Abramson2024} to atom-level
representations~\citep{jin2025oxtal}. MolCrystalFlow represents each molecule as a
rigid body and learns geodesic flows over molecular centroids, orientations, and
lattice parameters~\citep{zeng2026molcrystalflow}. PackFlow applies
reinforcement-learning post-training guided by MLIP energies and forces, improving
physical validity and concentrating proposals in low-energy
basins~\citep{subramanian2026packflow}. CLARI generates explicit unit cells and
replaces triangle layers with pair-bias attention, yielding a reported
15--30$\times$ speedup over OXtal~\citep{lo2026clari}. However, none of these
methods provides a single model in which molecular conformers, stereochemical
labels, and space-group information can each be supplied or omitted at inference
time~\citep{jin2025oxtal,lo2026clari,subramanian2026packflow,
zeng2026molcrystalflow}. The capabilities of our model and prior approaches are
summarized in \Cref{tab:model-capability-comparison}.

\begin{table*}[t]
\centering
\caption{\textbf{Capabilities and scope of generative molecular CSP models.}
We compare explicit hydrogen generation, chemical scope, and conditioning inputs.
\capsupported{} and \capunsupported{} denote supported and unsupported capabilities.
For local 3D input, \caprequired{}, \capsupported{}, and \capunsupported{} denote
required, optional, and unsupported conditioning, respectively.}
\label{tab:model-capability-comparison}
\small
\setlength{\tabcolsep}{4.0pt}
\begin{tabular}{@{}lcccccc@{}}
\toprule
& \multicolumn{1}{c}{Generation}
& \multicolumn{2}{c}{Chemical/data scope}
& \multicolumn{3}{c}{Conditioning inputs} \\
\cmidrule(lr){2-2}\cmidrule(lr){3-4}\cmidrule(l){5-7}
Method
& \shortstack{Explicit H\\output}
& \shortstack{Multi-\\component}
& \shortstack{Organo-\\metallics}
& \shortstack{Local 3D\\input}
& \shortstack{Stereo\\labels}
& \shortstack{Space\\group} \\
\midrule
OXtal~\citep{jin2025oxtal}
& \capunsupported{} & \capsupported{} & \capsupported{}
& \caprequired{} & \capunsupported{} & \capunsupported{} \\
PackFlow~\citep{subramanian2026packflow}
& \capunsupported{} & \capunsupported{} & \capunsupported{}
& \capunsupported{} & \capunsupported{} & \capunsupported{} \\
MolCrystalFlow~\citep{zeng2026molcrystalflow}
& \capsupported{} & \capunsupported{} & \capunsupported{}
& \caprequired{} & \capunsupported{} & \capunsupported{} \\
CLARI~\citep{lo2026clari}
& \capsupported{} & \capsupported{} & \capsupported{}
& \capunsupported{} & \capunsupported{} & \capunsupported{} \\
\rowcolor{mfTealFill}
\model{}
& \capsupported{} & \capsupported{} & \capsupported{}
& \capsupported{} & \capsupported{} & \capsupported{} \\
\bottomrule
\end{tabular}
\end{table*}

\textbf{Generative CSP.}
A related line of work studies crystal generation and structure prediction for
inorganic materials. CDVAE combines a variational autoencoder with diffusion-based
coordinate generation, whereas DiffCSP jointly diffuses fractional coordinates and
lattice conditioned on a given
composition~\citep{xie2022crystal,jiao2024crystal}. Subsequent approaches formulate
periodic crystal generation using flow matching, as in FlowMM and
CrystalFlow~\citep{millerflowmm,luo2025crystalflow}; stochastic interpolants, as in
OMatG~\citep{hollmeropen}; or Bayesian flows over periodic
variables~\citep{wu2025periodic}.

Generative CSP has also been extended to metal-organic frameworks (MOFs), 
where metal nodes and organic linkers can be treated as modular building blocks. 
MOFFlow~\citep{kim2025mofflow} and MOF-BFN~\citep{jiao2025mof} generate lattices 
and rigid-body poses of known building blocks using Riemannian and Bayesian flows, 
respectively, while MOFFlow-2~\citep{kim2025flexible} additionally generates building
blocks and models linker torsions. AtomMOF~\citep{kim2026atommof} removes the rigid-body
assumption by directly generating MOF and MOF--adsorbate structures at all-atom resolution.
\section{Method}
\label{sec:method}

\model{} predicts atomic coordinates and a unit cell from a molecular specification, a formula-unit count $Z$, and optional auxiliary information using a conditional coordinate--lattice flow. We describe the selected \model{} configuration below, with design choices justified by the controlled studies in \Cref{sec:recipe}.

\subsection{Problem Formulation}
\label{sec:method-problem}

\textbf{Crystal representation.}
We represent a crystal with $N$ atoms by atom types
$\mathbf{A}=(A_1,\ldots,A_N)\in\mathcal{A}^{N}$, Cartesian coordinates
$\mathbf{X}=(\mathbf{x}_1,\ldots,\mathbf{x}_N)^{\top}\in\mathbb{R}^{N\times3}$, and a
lattice cell $\mathbf{L}\in\mathbb{R}^{3\times3}$ whose rows are lattice
vectors. Integer translations along these lattice vectors define the corresponding
infinite periodic crystal.

Since the same lattice can be represented by different basis vectors and global
rotations, we canonicalize the cell in two steps. First, we Niggli-reduce~\citep{krivy1976unified,grosse2004numerically} each
crystal to obtain a reduced basis $\widetilde{\mathbf{L}}$, reducing the ambiguity among equivalent lattice bases. Second, we remove the remaining global rotational
freedom by forming the rotation-invariant Gram matrix
$\widetilde{\mathbf{L}}\widetilde{\mathbf{L}}^{\top}$ and taking its unique
lower-triangular Cholesky factor with positive diagonal:
\begin{equation*}
\mathbf{L}=\operatorname{chol}\!\left(
\widetilde{\mathbf{L}}\widetilde{\mathbf{L}}^{\top}
\right)=
\begin{pmatrix}
e^{\ell_1} & 0 & 0 \\
\ell_2 & e^{\ell_3} & 0 \\
\ell_4 & \ell_5 & e^{\ell_6}
\end{pmatrix}.
\end{equation*}
We further parameterize the canonical cell by an unconstrained vector  $\boldsymbol{\ell}=(\ell_1,\ldots,\ell_6)\in\mathbb{R}^{6}$ following \citet{veljkovic2026crystalite}.

\textbf{Molecular crystal structure prediction (CSP).}
We formulate molecular CSP as conditional generation of Cartesian coordinates
$\mathbf{X}$ and the lattice latent $\boldsymbol{\ell}$ from a chemical
specification, a supplied formula-unit count $Z$, and optional auxiliary
information. The chemical specification consists of $K$ distinct molecular
graphs $\mathcal{G}=\{G_k\}_{k=1}^{K}$ and a stoichiometry vector
$\mathbf{r}=(r_1,\ldots,r_K)$, where $r_k$ denotes the number of copies of
component $G_k$ in one formula unit. For example, a binary $2{:}1$ cocrystal
has $\mathbf{r}=(2,1)$. Each molecular graph specifies atom types, bond types,
and formal charges. To be specific, if component $G_k$ contains $N_k$ atoms, a unit cell with $Z$ formula units
contains
\[
N = Z\sum_{k=1}^{K} r_k N_k
\]
atoms. Optional information $\mathcal{O}$ may include molecular templates,
stereochemical annotations, and a space-group label. A molecular template is a
reference 3D conformer, generated for example with RDKit~\citep{rdkit}, that
provides local molecular geometry without specifying crystal packing. Given the
complete condition $\mathcal{C}=(\mathcal{G},\mathbf{r},Z,\mathcal{O})$, the
generator models $p_{\theta}(\mathbf{X},\boldsymbol{\ell}\mid\mathcal{C})$.

\subsection{Variational Flow Matching for Molecular CSP}
\label{sec:method-vfm}

We use variational flow matching \citep[VFM;][]{eijkelboom2024variational}, which generalizes flow matching~\citep{lipman2023flow}
through a flexible choice of the
variational posterior~\citep{zaghen2025riemannian}.

\textbf{Training objective.} Let $(\mathbf{y}_1,\mathcal{C})$ denote a crystal state and its condition, and let
$\mathbf{y}_0\sim p_0$ be a prior sample.
We use the linear conditional path to interpolate between the prior and data endpoints:
\begin{equation}
\label{eq:method-flow-path}
\mathbf{y}_t=(1-t)\mathbf{y}_0+t\mathbf{y}_1.
\end{equation}

Given a noisy state $\mathbf{y}_t$,
VFM learns a variational posterior
$q_{\theta}(\mathbf{y}_1\mid\mathbf{y}_t,t,\mathcal{C})$
that approximates the true posterior by minimizing the expected negative log-likelihood,
\begin{equation}
\label{eq:method-vfm-general}
\mathcal{L}_{\mathrm{VFM}}(\theta)
=-\mathbb{E}_{\mathbf{y}_1,\mathbf{y}_t,t}\!\left[
\log q_{\theta}(\mathbf{y}_1\mid\mathbf{y}_t,t,\mathcal{C})
\right].
\end{equation}
The variational family determines the form of this objective.
We model the posterior as a fully factorized Laplace distribution with a fixed scale,
which yields a component-weighted $L^1$ loss on the predicted endpoints.

The model parameterizes the mean of the variational posterior,
which is equivalent to predicting the clean endpoint $\mathbf{y}_1$ from the noisy
state $\mathbf{y}_t$ and condition $\mathcal{C}$ as
$\widehat{\mathbf{X}}_1,\widehat{\boldsymbol{\ell}}_1
=\boldsymbol{\mu}_t^{\theta}\!\left(\mathbf{y}_t,\mathcal{C}\right)$.
The equation then simplifies to
\begin{equation}
\label{eq:method-vfm-loss}
\mathcal{L}_{\mathrm{VFM}}(\theta)
=\mathbb{E}_{\mathbf{y}_1,\mathbf{y}_t,t}\!\left[
\frac{\lambda_{\mathrm{coord}}}{3N}
\left\|\widehat{\mathbf{X}}_1-\mathbf{X}_1\right\|_1
+
\frac{\lambda_{\mathrm{lattice}}}{6}
\left\|\widehat{\boldsymbol{\ell}}_1-
\boldsymbol{\ell}_1\right\|_1
\right],
\end{equation}
where $\lambda_{\mathrm{coord}}$ and $\lambda_{\mathrm{lattice}}$ are loss weights
for the coordinate and lattice components, respectively.

\textbf{Sampling.}
After training $\boldsymbol{\mu}_t^\theta$, we generate samples by solving the ordinary differential equation (ODE) or the corresponding stochastic differential equation (SDE)~\citep{eijkelboom2024variational,albergo2025stochastic}:
\begin{equation}
d\mathbf{y}_t
=\mathbf{v}_t^\theta(\mathbf{y}_t)\,dt,
\qquad
d\mathbf{y}_t
=
\left[
\mathbf{v}_t^\theta(\mathbf{y}_t)
+\frac{g_t^2}{2}\mathbf{s}_t^\theta(\mathbf{y}_t)
\right]dt
+g_t\,d\mathbf{W}_t.
\end{equation}
For the linear path,
$\mathbf{v}_t^\theta(\mathbf{y}_t)
=(\boldsymbol{\mu}_t^\theta(\mathbf{y}_t,\mathcal C)-\mathbf{y}_t)/(1-t)$
and
$\mathbf{s}_t^\theta(\mathbf{y}_t)
=(t\,\mathbf{v}_t^\theta(\mathbf{y}_t)-\mathbf{y}_t)/(1-t)$.
Here, $g_t$ is the diffusion coefficient controlling the stochasticity, and
$\mathbf{W}_t$ is a standard Wiener process.

\subsection{Model Architecture}
\label{sec:method-architecture}

Our model $\boldsymbol{\mu}_t^{\theta}$ mainly employs a Pairmixer-based
architecture~\citep{ouyangzhang2025triangle} with trunk-head modularization~\citep{Abramson2024, bytedance2025protenix}. As shown in
\Cref{fig:method-architecture}, the model first constructs condition-only
single and pair representations, refines the pair representation with Pairmixer,
injects the noisy crystal state into the single representation, and processes it
with a Diffusion Transformer (DiT)~\citep{peebles2023scalable}. Separate output heads predict the coordinate and lattice endpoints. The complete architecture is specified
algorithmically in \Cref{app:model-architecture}.

\input{figures/fig_method_architecture.tex}

\textbf{Condition embedding.}
The input embedder embeds the condition
$\mathcal{C}=(\mathcal{G},\mathbf{r},Z,\mathcal{O})$ into a per-atom single
representation $\mathbf{S}_{\mathcal C}$ and an initial per-atom-pair
representation $\mathbf{P}^{(0)}$:
\begin{equation*}
\mathbf{S}_{\mathcal C}=E_{\mathrm{single}}(\mathcal C),
\qquad
\mathbf{P}^{(0)}=E_{\mathrm{pair}}(\mathbf{S}_{\mathcal C},\mathcal C),
\end{equation*}
where $E_{\mathrm{single}}$ and $E_{\mathrm{pair}}$ are learned single and pair
embedding networks. The single representation combines atom identity,
periodic-table descriptors, formal charge, optional template coordinates,
chirality, and space group. The pair representation combines projected single
features with intramolecular bond types, bond stereochemistry, template
displacements, and template distances. Availability masks and learned null states
distinguish unavailable optional conditions from provided values.

\textbf{Pairmixer.}
Pairmixer is a streamlined alternative to Pairformer that retains incoming and
outgoing triangle multiplication and pair transitions while omitting triangle
attention. Each block updates the current pair representation $\mathbf{P}$ through
sequential residual operations,
\begin{equation*}
\mathbf{P}\leftarrow\mathbf{P}
+\operatorname{TriMul}_{\mathrm{out}}(\mathbf{P}),\qquad
\mathbf{P}\leftarrow\mathbf{P}
+\operatorname{TriMul}_{\mathrm{in}}(\mathbf{P}),\qquad
\mathbf{P}\leftarrow\mathbf{P}
+\operatorname{Transition}(\mathbf{P}),
\end{equation*}
where $\operatorname{TriMul}_{\mathrm{out}}$ and
$\operatorname{TriMul}_{\mathrm{in}}$ denote outgoing and incoming triangle
multiplication, respectively. The transition is a gated feed-forward network
applied independently to each atom pair. Pairmixer leaves the single representation unchanged and produces a refined pair representation $\mathbf{P}^{\star}$.

\textbf{Noisy-state injection.}
The noisy crystal state is injected into the single representation
$\mathbf{S}_{\mathcal C}$. Coordinate features are encoded using Fourier features~\citep{tancik2020fourier}, while the lattice is embedded
globally and broadcast to all atoms. The resulting single representation is
\begin{equation*}
\mathbf{S}_t
=\mathbf{S}_{\mathcal C}
+E_X(\mathbf{X}_t)
+\operatorname{Broadcast}\!\left(E_L(\boldsymbol{\ell}_t)\right),
\end{equation*}
where $E_X$ and $E_L$ denote the coordinate and lattice embedding networks,
respectively. A sinusoidal embedding of the flow time $t$ provides a separate
global condition to the transformer.

\textbf{DiT trunk.}
The trunk follows the DiT architecture, alternating between self-attention and
feed-forward updates to the single representation. In each attention layer, the
refined pair representation $\mathbf{P}^{\star}$ is projected to a
head-specific bias and added to the attention logits:
\begin{equation*}
a_{ij}^{(h)}
=
\frac{\left\langle\mathbf{q}_i^{(h)},\mathbf{k}_j^{(h)}\right\rangle}
{\sqrt{d_h}}
+b_h\!\left(\mathbf{P}_{ij}^{\star}\right),
\end{equation*}
where $\mathbf{q}_i^{(h)}$ and $\mathbf{k}_j^{(h)}$ are the query and key
vectors for atoms $i$ and $j$ in attention head $h$, $d_h$ is the head
dimension, and $b_h$ maps the pair representation to a scalar attention bias.
The time embedding modulates the normalization and residual gates of both
self-attention and feed-forward updates~\citep{peebles2023scalable}.

\textbf{Prediction heads.}
The DiT output $\mathbf{S}^{\mathrm{out}}$ is decoded by separate coordinate and
lattice endpoint heads. For the set $\mathcal V$ of valid atoms, the coordinate
head predicts an atomwise vector $\mathbf r_i$ and subtracts its mean over valid
atoms, while the lattice head applies an MLP to the mean-pooled single
representation:
\begin{equation*}
\mathbf{r}_i
= W_X \operatorname{Norm}(\mathbf{S}^{\mathrm{out}}_i),\qquad
\widehat{\mathbf{X}}_{1,i}
= \mathbf{r}_i
- \frac{1}{|\mathcal V|}\sum_{j\in\mathcal V}\mathbf{r}_j,\qquad
\widehat{\boldsymbol{\ell}}_1
= \operatorname{MLP}\!\left(
\frac{1}{|\mathcal V|}\sum_{i\in\mathcal V}
\operatorname{Norm}(\mathbf{S}^{\mathrm{out}}_i)
\right),
\end{equation*}
where $W_X$ is the coordinate projection and
$\widehat{\mathbf{X}}_1$ and $\widehat{\boldsymbol{\ell}}_1$ denote the predicted
coordinate and lattice endpoints.
\section{Benchmark Evaluation}
\label{sec:benchmarks}

Inspired by the CSP blind tests~\citep{hunnisett2024seventhranking,hunnisett2024seventhgeneration}, we evaluate models along two complementary axes: structure generation and structure ranking. Structure generation (\Cref{sec:structure-generation}) measures unranked coverage at a fixed candidate budget, testing whether a generator can recover the experimental structure. Structure ranking (\Cref{sec:structure-ranking}) measures whether generated candidates remain useful after downstream relaxation and ranking, by evaluating how highly the experimental form is ranked. Unlike the CSP blind-test ranking track, which fixes the candidate pool and varies the ranking method, we fix the downstream evaluator and vary the generator, thereby measuring end-to-end generator--evaluator compatibility.

\subsection{Experimental Setup}
\label{sec:setup}

\textbf{Dataset.}
We use the official CLARI training and validation split~\citep{lo2026clari}, derived from the Cambridge Structural Database (CSD)~\citep{groom2016cambridge}. The held-out test set comprises the OXtal Rigid and Flexible benchmarks~\citep{jin2025oxtal}, CSP5~\citep{bardwell2011fifth}, CSP6~\citep{reilly2016sixth}, CSP7~\citep{hunnisett2024seventhranking,hunnisett2024seventhgeneration}, and the CSD Teaching Subset~\citep{battle2010applications,lo2026clari}. To prevent leakage, CLARI excludes all entries from test refcode families and training structures that share an RDKit-sanitizable molecular component with more than seven heavy atoms, then holds out 1{,}000 refcode families for validation.

CLARI limits training and validation unit cells to 512 atoms, yielding 917{,}014 training and 1{,}048 validation entries. Because we add missing hydrogens with the CSD Python API~\citep{sykes2024scripting} before counting atoms, we reapply the 512-atom limit after hydrogen completion, leaving 912{,}807 training and 1{,}047 validation entries. \Cref{app:dataset-statistics} provides detailed dataset statistics.

\textbf{Training and sampling configuration.}
We train \ours{}-M (88M parameters) and \ours{}-L (187M parameters) using
Muon~\citep{liu2025muon} for hidden matrix parameters and
AdamW~\citep{loshchilov2019decoupled} for all remaining parameters. Both models
use a learning rate of $10^{-4}$, an effective batch size of 128, and an EMA
decay of $0.9999$; weight decay is 0 for \ours{}-M and $10^{-2}$ for
\ours{}-L. We also use bfloat16 and train on eight NVIDIA H200 GPUs.

At inference, we condition on stereochemistry when available and on RDKit-generated molecular conformers, reflecting information that is either specified in CSP blind
tests~\citep{reilly2016sixth} or readily derived from molecular
graphs~\citep{rdkit}. We do not use space-group conditioning because space groups
are generally not provided in CSP blind tests.
We generate candidates with the EDM--Heun sampler~\citep{karras2022elucidating},
using stochastic churn, $\rho=7$, and 200 sampling steps. Complete
hyperparameters are provided in
\Cref{app:training-sampling-hyperparameters}.

\textbf{Two-stage training.}
To enable faster iteration before scaling to larger structures, we follow the two-stage training of SeedFold~\citep{yi2025seedfold}. In the first stage, we train on crystals with at most 300 atoms; in the second, we expand to all crystals with at most 512 atoms. We save checkpoints every 50 epochs and select the model with the best validation $\mathrm{Sol}_C$.

\subsection{Structure Generation}
\label{sec:structure-generation}

We first evaluate structure generation alone, following the structure generation phase of the seventh CCDC CSP blind test~\citep{hunnisett2024seventhgeneration}. We measure whether each generative model can recover the experimental structure within a fixed candidate budget, without energy evaluation, relaxation, or ranking.

\textbf{Benchmarks.} We evaluate on six benchmarks: the OXtal Rigid and Flexible sets, with 50 targets each~\citep{jin2025oxtal}; CSP5, CSP6, and CSP7, with 6, 5, and 8 targets, respectively~\citep{bardwell2011fifth,reilly2016sixth,hunnisett2024seventhgeneration}; and CLARI's CSD Teaching Subset, with 773 targets~\citep{battle2010applications,lo2026clari}.

\textbf{Metric and protocol.} Following OXtal~\citep{jin2025oxtal} and CLARI~\citep{lo2026clari}, we report the crystal-level approximate solve rate $\mathrm{Sol}_C$ at candidate budgets of 30 and 1{,}000. A target is considered solved if any candidate is collision-free, at least 8 of 15 molecules are matched, and $\mathrm{RMSD}_{15}<2\text{\AA}$ with COMPACK~\citep{chisholm2005compack}. $\mathrm{Sol}_C$ is the fraction of solved targets. We also report the stricter $\mathrm{Sol}_C^{15/15}$, which requires all 15 molecules to match. For the 30-candidate budget, we follow CLARI's bootstrapping protocol~\citep{lo2026clari}: from each fixed pool of 1{,}000 candidates, we draw 5{,}000 resamples of 30 candidates and average the resulting $\mathrm{Sol}_C$. Full metric definitions are provided in \Cref{app:structure-generation-evaluation}.

\textbf{Baselines.} We compare against OXtal and three CLARI variants: CLARI-M, CLARI-L, and CLARI-H~\citep{jin2025oxtal,lo2026clari}. For a consistent comparison, we re-evaluate the official CLARI checkpoints using the OXtal evaluation protocol, since the two evaluators differ in collision detection and reference-structure selection. See \Cref{app:structure-generation-evaluation} for details.

\begin{figure}[t]
  \centering
  \includegraphics[width=\textwidth]{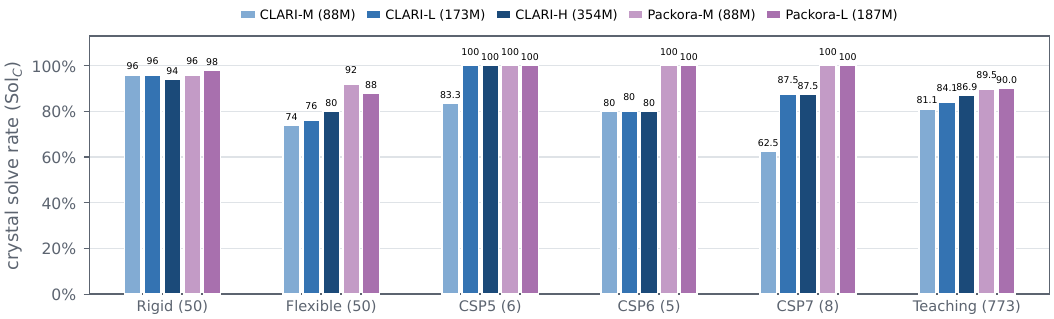}
    \caption{\textbf{Crystal coverage with 1{,}000 candidates per target.}
    We report the crystal-level solve rate $\mathrm{Sol}_C$ at a generation budget of
    1{,}000 candidates per target. Numbers in parentheses indicate the number of targets in
    each benchmark. \model{} achieves the highest solve rate on all six benchmarks.}
  \label{fig:benchmark-exact}
\end{figure}

\begin{figure}[t]
  \centering
  \includegraphics[width=\textwidth]{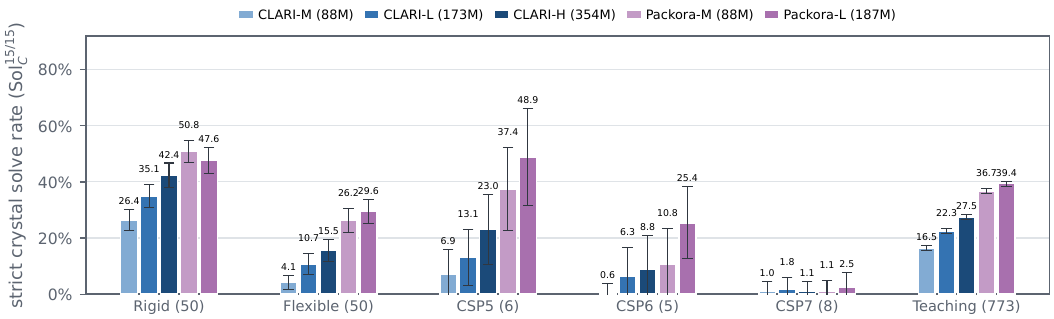}
  \caption{\textbf{Crystal coverage with 30 candidates per target under the strict criterion.} We report the crystal-level solve rate $\mathrm{Sol}_C^{15/15}$ at a generation budget of 30 candidates per target. $\mathrm{Sol}_C^{15/15}$ requires all 15
  molecules to match using COMPACK~\citep{bardwell2011fifth}. Numbers in
  parentheses indicate the number of targets in each benchmark. We generate
  1{,}000 candidates per target and report the mean over 5{,}000 bootstrap
  resamples of 30 candidates; error bars show one standard deviation across
  resamples. \model{} achieves the highest solve rate on all six benchmarks.}
  \label{fig:benchmark-teaser-strict}
\end{figure}

\begin{figure}[t]
    \centering
    \includegraphics[width=\textwidth]{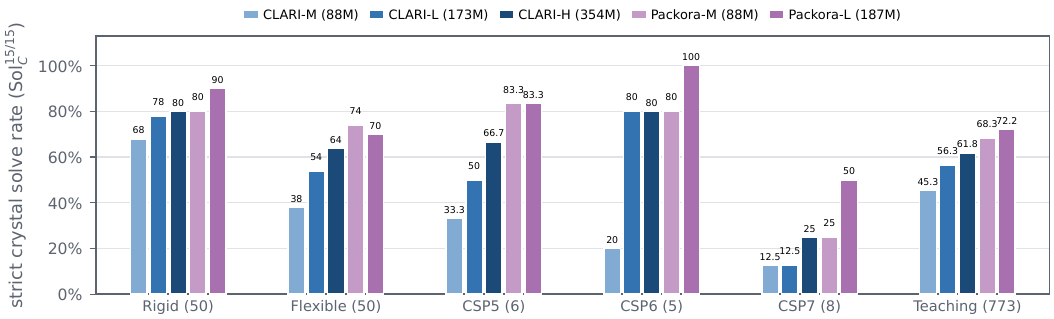}
    \caption{\textbf{Crystal coverage with 1{,}000 candidates per target under the strict criterion.} We report the crystal-level solve rate $\mathrm{Sol}_C^{15/15}$ at a generation budget of 1{,}000 candidates per target. $\mathrm{Sol}_C^{15/15}$ requires all 15 molecules to match under COMPACK~\citep{bardwell2011fifth}. Numbers in parentheses indicate the number of targets in each benchmark. \model{} achieves the highest solve rate on all six benchmarks.}
    \label{fig:benchmark-exact-strict}
\end{figure}

\textbf{Results.} Across all six benchmarks, \model{} achieves the highest solve rate at both 30- and 1{,}000-candidate budgets under the standard
and strict criteria (\Cref{fig:benchmark-teaser,fig:benchmark-exact,fig:benchmark-teaser-strict,fig:benchmark-exact-strict}),
outperforming the strongest baseline in 23 of 24 settings and tying on the saturated CSP5 result. At 30 candidates, the relative gain
reaches 54.5\% on CSP7 under the standard criterion and 188.0\% on CSP6 under the strict criterion.
\Cref{fig:strict-structure-examples} further demonstrates accurate recovery across diverse molecules and packing motifs.
Full tables are provided in \Cref{app:structure-generation-results}, results with $Z$ sampled from an empirical prior in
\Cref{app:empirical-unit-cell-z}, and aligned structure overlays in \Cref{app:strict-structure-overlays}.

\subsection{Structure Ranking}
\label{sec:structure-ranking}

\textbf{Benchmarks.}
We evaluate two benchmarks derived from FastCSP~\citep{gharakhanyan2025fastcsp}. The single-polymorph benchmark contains 23 semi-rigid and three flexible systems, each with one experimental form. The multi-polymorph benchmark contains five semi-rigid and three flexible systems, comprising 29 experimental forms from 28 distinct CSD entries. We exclude BEDMIG, UNOGIN, BISMEV, and BIYSEH from the multi-polymorph benchmark because their CSD families appear in the CLARI training split.

\textbf{Relaxation and ranking workflow.} 
For each CSD entry, each generator produces 1{,}000 independent candidates. We generally follow the FastCSP post-generation workflow, relaxing candidates using UMA-S-1.2 with the OMC task~\citep{wood2025family} and the ASE BFGS optimizer~\citep{larsen2017atomic}. We use a maximum-force threshold of $0.02$~eV,\AA$^{-1}$ and up to 500 optimization steps. We discard candidates that fail to converge, change molecular connectivity or $Z$, or have densities outside $0.5$--$3.0$~g,cm$^{-3}$. We then retain structures within 10~kJ,mol$^{-1}$ of the minimum energy and deduplicate equivalent relaxed structures. For each multi-polymorph system, we pool candidates generated from all constituent CSD entries before energy filtering, deduplication, and ranking, so that all experimental forms are evaluated within a shared energy ordering.

\textbf{Metrics.}
We rank the remaining unique candidates by increasing lattice
energy~\citep{gharakhanyan2025fastcsp}. For candidate
$\mathcal{S}_i=(\mathbf{A}_i,\mathbf{X}_i,\mathbf{L}_i)$, we compute the
lattice energy per formula unit as
\begin{equation}
  E_{\mathrm{latt}}(\mathcal{S}_i)
  = \frac{E_{\mathrm{UMA}}(\mathcal{S}_i)}{Z_i}
  - E_{\mathrm{mol}},
\end{equation}
where $Z_i$ is the number of formula units in the cell and $E_{\mathrm{mol}}$
is the isolated-component reference energy for one formula unit.

A candidate matches an experimental CSD form under
COMPACK~\citep{chisholm2005compack} if 30 molecules match with
$\mathrm{RMSD}_{30}<1$~\AA. For each experimental form, we record the best
energy rank among all matching candidates. We report overall recovery at any rank, Top-$k$ recovery for $k\in{1,5,20}$, and the mean best rank among recovered forms.

\textbf{Baseline.}
We compare \model{}-M and \model{}-L with CLARI-H. We do not include FastCSP as
a baseline because its random structure generator uses a
substantially larger search budget: approximately 53{,}000--300{,}000 raw
structures per target, with 10{,}000--176{,}000 retained for relaxation after
deduplication, compared with 1{,}000 candidates per CSD entry in our
evaluation~\citep{gharakhanyan2025fastcsp}.

\begin{figure}[t]
    \centering
    \includegraphics[width=\textwidth]{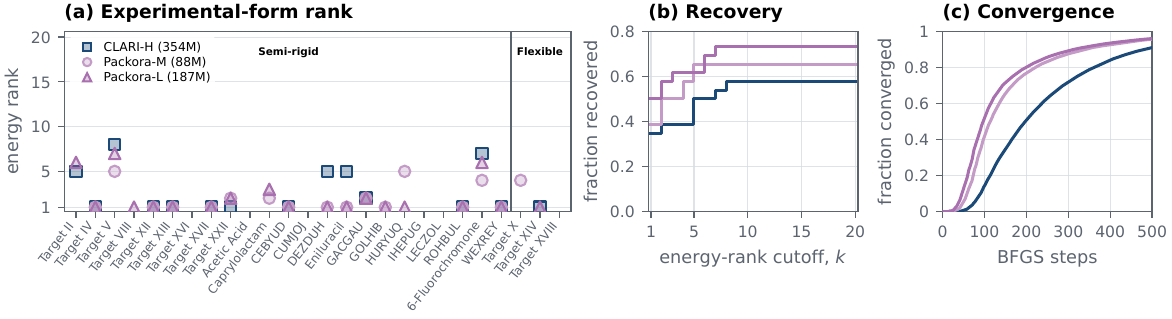}
    \caption{\textbf{FastCSP single-polymorph benchmark.}
    We generate 1{,}000 candidates for each of the 26 targets, relax them with
    UMA-S-1.2~\citep{wood2025family}, and rank them by increasing energy.
    \textbf{(a)} Best energy rank of the experimental target form for CLARI-H,
    \model{}-M, and \model{}-L; lower is better, with rank 1 denoting the predicted
    minimum-energy structure. \model{}-L and \model{}-M recover 19 and 17 targets
    within the top 20, respectively, compared with 15 for CLARI-H.
    \textbf{(b)} Fraction of targets recovered by energy-rank cutoff $k$; \model{}
    achieves higher recovery across cutoffs. \textbf{(c)} Cumulative fraction of the
    generated candidates converged by BFGS step; \model{} candidates converge faster
    throughout relaxation. Relaxing all candidates on eight NVIDIA H200 GPUs takes
    16.8, 11.4, and 10.2~h for CLARI-H, \model{}-M, and \model{}-L, respectively.}
    \label{fig:fastcsp-single-ranking}
\end{figure}

\textbf{Single-polymorph results.}
\model{}-M and \model{}-L recover 17 and 19 of the 26 targets, respectively,
compared with 16 for CLARI-H, while reducing the mean rank among recovered
targets from 4.25 to 2.00 and 2.05 (\Cref{tab:fastcsp-single-ranking} and \Cref{fig:fastcsp-single-ranking}). The recovery advantage persists across all evaluated rank cutoffs. \model{} candidates also relax faster: \model{}-L reaches 50\% convergence after 98 BFGS steps, compared with 198 for CLARI-H. On eight NVIDIA H200 GPUs, relaxing all candidates takes
11.4~h for \model{}-M and 10.2~h for \model{}-L, compared with 16.8~h for CLARI-H. Per-target ranks and $\mathrm{RMSD}_{30}$ values are reported in \Cref{app:structure-ranking-results}.

\begin{table}[t]
\centering
\caption{\textbf{Experimental-form recovery on FastCSP single-polymorph benchmark.}
We generate 1{,}000 candidates for each of the 26 targets, relax them with
UMA-S-1.2~\citep{wood2025family}, and rank the relaxed structures by increasing
energy. Recovered denotes the fraction of targets with an
experimental-form match at any rank, while Top-$k$ denotes the fraction whose
best match appears within the top $k$ ranks. Mean rank is computed over recovered
targets only, with lower values indicating better ranking. \textbf{Bold} marks
the best result. \model{} achieves both higher recovery across rank cutoffs and
lower experimental-form ranks.}
\label{tab:fastcsp-single-ranking}
\begin{tabular}{lccccc}
\toprule
Method & Recovered $\uparrow$ & Top-1 $\uparrow$ & Top-5 $\uparrow$
& Top-20 $\uparrow$ & Mean rank $\downarrow$ \\
\midrule
CLARI-H & 16/26 & 9/26 & 13/26 & 15/26 & 4.25 \\
\rowcolor{mfTealFill}
\model{}-M & 17/26 & 10/26 & \textbf{17/26} & 17/26 & \textbf{2.00} \\
\rowcolor{mfTealFill}
\model{}-L & \textbf{19/26} & \textbf{13/26} & 16/26 & \textbf{19/26} & 2.05 \\
\bottomrule
\end{tabular}
\end{table}

\begin{figure}[t]
  \centering
  \includegraphics[width=\textwidth]{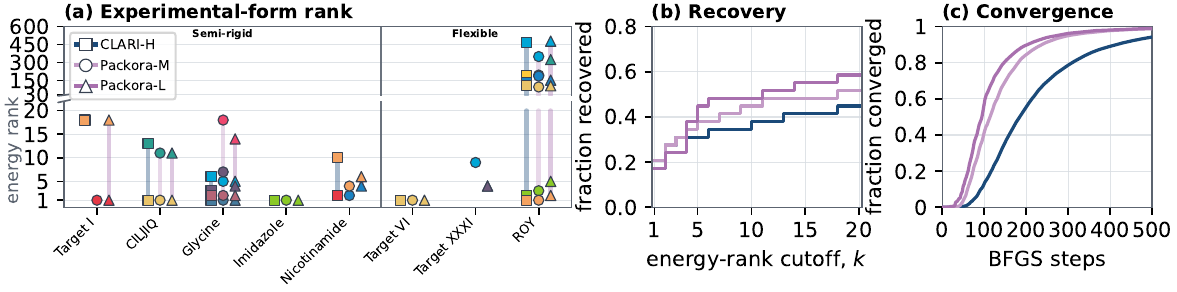}
  \caption{\textbf{FastCSP multi-polymorph benchmark.}
    We generate 1{,}000 candidates for each of the 29 target structures
    across eight systems, relax them with
    UMA-S-1.2~\citep{wood2025family}, and rank them by increasing energy within
    each system. \textbf{(a)} Best energy rank of each of the 29 polymorphs
    for CLARI-H, \model{}-M, and \model{}-L; lower is better, with rank
    1 denoting the predicted minimum-energy structure. Marker color identifies
    the polymorph across systems. \model{}-L and \model{}-M recover
    21 and 19 polymorphs, respectively,
    compared with 18 for CLARI-H. \textbf{(b)} Fraction of
    polymorphs recovered by energy-rank cutoff $k$; \model{} achieves higher
    recovery across cutoffs. \textbf{(c)} Cumulative fraction of the generated candidates converged by BFGS step; \model{} candidates converge faster throughout relaxation.
    Relaxing all candidates on eight NVIDIA H200 GPUs takes 16.8, 10.6, and
    8.9~h for CLARI-H, \model{}-M, and \model{}-L, respectively.}
  \label{fig:fastcsp-multi-ranking}
\end{figure}

\textbf{Multi-polymorph results.}
The advantage remains in the multi-polymorph setting. \model{}-M and \model{}-L
recover 19 and 21 of the 29 forms, respectively, compared with 18 for CLARI-H,
and reduce the mean recovered rank from 102.17 to 47.11 and 54.62
(\Cref{tab:fastcsp-multi-ranking,fig:fastcsp-multi-ranking}). Also,
both \model{} variants recover at least one polymorph for all 8 of 8 systems,
whereas CLARI-H recovers 7. Relaxation is also faster: after 100 BFGS steps,
41.4\% and 57.3\% of \model{}-M and \model{}-L candidates have converged,
compared with 13.0\% for CLARI-H. Total relaxation takes 10.6~h and 8.9~h,
respectively, versus 16.8~h for CLARI-H on eight NVIDIA H200 GPUs. Per-form
ranks and $\mathrm{RMSD}_{30}$ values are reported in
\Cref{app:structure-ranking-results}.

\begin{table}[t]
\centering
\caption{\textbf{Experimental-form recovery on FastCSP multiple-polymorph benchmark.}
We generate 1{,}000 candidates for each of the 29 targets, relax them with
UMA-S-1.2~\citep{wood2025family}, and rank the relaxed structures by increasing
energy. Recovered denotes the fraction of targets with an
experimental-form match at any rank, while Top-$k$ denotes the fraction whose
best match appears within the top $k$ ranks. Mean rank is computed over recovered
targets only, with lower values indicating better ranking. \textbf{Bold} marks
the best result. \model{} achieves both higher recovery across rank cutoffs and
lower experimental-form ranks.}
\label{tab:fastcsp-multi-ranking}
\begin{tabular}{lccccc}
\toprule
Method & Recovered $\uparrow$ & Top-1 $\uparrow$ & Top-5 $\uparrow$
& Top-20 $\uparrow$ & Mean rank $\downarrow$ \\
\midrule
CLARI-H & 18/29 & 5/29 & 9/29 & 13/29 & 102.17 \\
\rowcolor{mfTealFill}
\model{}-M & 19/29 & \textbf{6/29} & 11/29 & 15/29 & \textbf{47.11} \\
\rowcolor{mfTealFill}
\model{}-L & \textbf{21/29} & 5/29 & \textbf{13/29} & \textbf{17/29} & 54.62 \\
\bottomrule
\end{tabular}
\end{table}

\section{Building the Recipe}
\label{sec:recipe}
\label{sec:ablations}

\begin{figure}[!t]
  \centering
  \includegraphics[width=\textwidth]{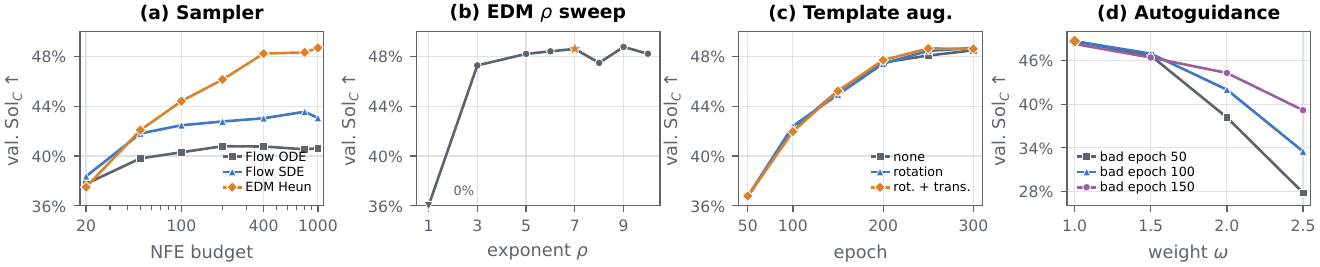}
  \caption{\textbf{Inference ablations.}
    \textbf{(a)} Validation $\mathrm{Sol}_C$ versus NFE budget for flow ODE,
    flow SDE, and EDM--Heun. \textbf{(b)} EDM--Heun with 400 NFEs across
    discretization exponents $\rho$; the failed $\rho=1$ result is shown at the
    lower boundary, and the star marks the selected $\rho=7$.
    \textbf{(c)} Validation trajectories with no template augmentation,
    rotation only, and rotation plus translation at inference.
    \textbf{(d)} Autoguidance using degraded checkpoints from epochs 50, 100,
    and 150 across guidance weights $\omega$.}
  \label{fig:inference-sweeps}
\end{figure}

We construct the final \model{} recipe through five controlled studies of the
design space:
\begin{itemize}
    \item \textbf{Architecture:} Which backbone---DiT with pairwise bias,
    Pairformer, or PairMixer---is most effective? Where should the noisy crystal
    state enter the network, and do additional single-track or geometry modules
    help? (\Cref{fig:sp-architecture,fig:architecture-ablation-convergence})
    
    \item \textbf{Training:} Which choices of time distribution, augmentation,
    endpoint objective, loss weighting, auxiliary supervision, and optimizer
    improve structure generation?
    (\Cref{fig:training-recipe-summary})
    
    \item \textbf{Conditioning:} How should optional inputs be dropped during
    training so that a single model remains effective across different
    combinations of stereochemistry, molecular templates, and space-group
    information? (\Cref{fig:conditioning-sweep})
    
    \item \textbf{Inference:} Which sampler and solver settings work best, and
    does autoguidance provide further gains?
    (\Cref{fig:inference-sweeps})

    \item \textbf{Scaling:} How does scaling single-track and pair-track widths
    affect per-sample success and coverage across targets, and which allocation
    best balances the two?
    (\Cref{fig:width-scaling-tradeoff};
    \Cref{tab:width_scaling_ablation})
\end{itemize}

\subsection{Ablation Protocol}
\label{sec:ablations-setup}

\textbf{Dataset preprocessing.}
Following OXtal~\citep{jin2025oxtal}, we construct the dataset from CSD entries
deposited by May 1, 2025. We require three-dimensional coordinates, an
$R$-factor of at most $9\%$, single-crystal X-ray diffraction at ambient
pressure, a non-polymeric structure, and a known space group. We add missing
hydrogens using the CSD API~\citep{sykes2024scripting} and retain unit cells
with at most 512 atoms, including hydrogens. To prevent benchmark leakage, we
remove all entries belonging to a test CSD family or containing an eligible
CSD-provided component SMILES found in the test sets. Within each remaining CSD
family, we deduplicate equivalent structures using pymatgen
StructureMatcher~\citep{ong2013pymatgen} with $\operatorname{ltol}=0.2$,
$\operatorname{stol}=0.3$, and $\operatorname{angle\_tol}=5^\circ$, retaining the entry
with the lowest $R$-factor. The complete preprocessing pipeline and dataset
statistics are provided in \Cref{app:dataset-preprocessing}.

\textbf{Setup.}
To make controlled ablations feasible under a fixed compute budget, we train each
configuration for 300 epochs on structures with at most 300 atoms per unit cell,
saving checkpoints every 50 epochs. We evaluate each checkpoint on a fixed subset
of 8{,}192 validation crystals, generating 30 candidates per target with the
200-step EDM--Heun sampler, and report $\mathrm{Sol}_C$ as defined in
\Cref{sec:structure-generation}. Architecture, training, conditioning, and scaling
studies follow this protocol; inference studies instead vary the sampler and
function-evaluation budget. Within each study, we vary one design choice at a time.

\subsection{Architecture: A Cacheable Pairmixer Design}
\label{sec:recipe-architecture}

\textbf{Explored designs.} We compare three backbones: (1)~DiT with pair-bias attention, as used by CLARI~\citep{lo2026clari}; (2)~Pairformer, as used by AlphaFold3 and OXtal~\citep{Abramson2024,jin2025oxtal}; and (3)~Pairmixer, a lighter Pairformer variant without triangle attention~\citep{ouyangzhang2025triangle}. We also test a single-track update and the geometric enhancement module (GEM)~\citep{veljkovic2026crystalite}, which injects periodic minimum-image pair geometry from the noisy structure. Finally, we compare a pre-entry design, inspired by Proteina~\citep{geffner2025proteina}, where noisy coordinate and lattice features enter before pair refinement, with a post-entry design, following AlphaFold3~\citep{Abramson2024}, where they enter afterward so that the pair representation can be cached.

\usetikzlibrary{arrows.meta,positioning,backgrounds,fit,calc,decorations.pathreplacing}

\definecolor{archInputBg}{HTML}{E9F0F5}
\definecolor{archInputBlock}{HTML}{AFC9DA}
\definecolor{archInputLine}{HTML}{5D8198}
\definecolor{archTimeBg}{HTML}{FCF5E6}
\definecolor{archTimeBlock}{HTML}{F6DFB1}
\definecolor{archTimeLine}{HTML}{B7892D}
\definecolor{archPairBg}{HTML}{E3EEE9}
\definecolor{archPairBlock}{HTML}{C5DED1}
\definecolor{archPairLine}{HTML}{5F8D7A}
\definecolor{archTrunkBg}{HTML}{E7DCE3}
\definecolor{archHeadBg}{HTML}{F8EBE9}
\definecolor{archAttnBlock}{HTML}{EAC8B1}
\definecolor{archTriAttnBlock}{HTML}{F2D39B}
\definecolor{archMlpBlock}{HTML}{CDB6C5}
\definecolor{archArrow}{HTML}{2D2D2D}
\definecolor{archFrame}{HTML}{30343B}

\colorlet{cEmbed}{archInputBlock}
\colorlet{cEmbedDk}{archInputLine}
\colorlet{cEmbedBg}{archInputBg}
\colorlet{cAttn}{archAttnBlock}
\colorlet{cMLP}{archMlpBlock}
\colorlet{cTri}{archPairBlock}
\colorlet{cPairBg}{archPairBg}
\colorlet{cTrunkBg}{archTrunkBg}
\colorlet{cHeadBg}{archHeadBg}
\colorlet{cTime}{archTimeBlock}
\colorlet{cTimeBg}{archTimeBg}

\tikzset{
  archfig/.style={
    font=\scriptsize,
    >={Stealth[length=1.5mm]},
    box/.style={draw, line width=0.45pt, rounded corners=1.3pt, align=center, minimum height=6mm,
                inner sep=1.6pt, font=\scriptsize},
    emb/.style={box, draw=cEmbedDk, fill=cEmbed, minimum width=15mm},
    attn/.style={box, draw=archFrame, fill=cAttn, minimum width=13mm},
    triattn/.style={box, dashed, draw=archFrame, fill=archTriAttnBlock,
                    minimum width=15mm},
    trans/.style={box, draw=archFrame, fill=cMLP, minimum width=12mm},
    tri/.style={box, draw=archPairLine, fill=cTri, minimum width=13mm},
    time/.style={box, draw=archTimeLine, fill=cTime, minimum width=15mm},
    trunk/.style={box, draw=archFrame, fill=cTrunkBg, minimum width=29mm,
                  minimum height=15mm, font=\bfseries\scriptsize},
    heads/.style={box, draw=archFrame, fill=cHeadBg, minimum width=22mm,
                  minimum height=15mm, font=\bfseries\scriptsize},
    sum/.style={draw, line width=0.45pt, circle, inner sep=0pt, minimum size=3.2mm,
                font=\scriptsize},
    dot/.style={circle, fill=archArrow, inner sep=0pt, minimum size=2.2pt},
    fl/.style={->, draw=archArrow, line width=0.45pt},
    wire/.style={draw=archArrow, line width=0.45pt},
    lab/.style={font=\small, text=archFrame, inner sep=1pt},
    modbg/.style={rounded corners=4pt, draw, line width=0.45pt, inner sep=5pt},
    mtitle/.style={font=\bfseries\footnotesize},
  }
}

\newcommand{\skipdy}{5.5mm}

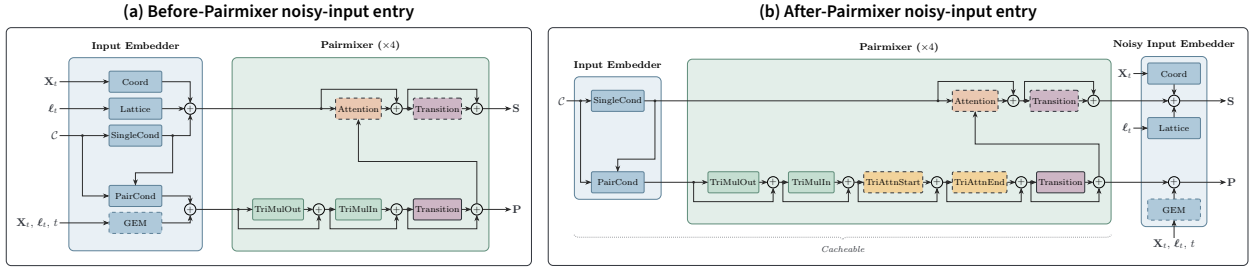
\begin{figure}[!t]
  \centering
  \resizebox{\textwidth}{!}{%
  \begin{minipage}{356mm}
    \centering
  \begin{minipage}[t]{150mm}
    \centering
    {\sffamily\bfseries\Large (a) Before-Pairmixer noisy-input entry}\\[1.5mm]
    \begin{tikzpicture}[archfig]
      \node[emb] (ce) {Coord};
      \node[emb, below=1.5mm of ce] (le) {Lattice};
      \node[emb, below=1.5mm of le] (sc) {SingleCond};
      \node[sum] (ssum) at ($(le.east)+(8mm,0)$) {$+$};
      \draw[fl] (ce.east) -| (ssum.north);
      \draw[fl] (le.east) -- (ssum);
      \draw[fl] (sc.east) -| (ssum.south);

      \node[emb, below=11mm of sc] (pc) {PairCond};
      \node[emb, dashed, below=1.5mm of pc] (ge) {GEM};
      \node[sum] (zsum) at ($(pc.east)!0.5!(ge.east)+(8mm,0)$) {$+$};
      \draw[fl] (pc.east) -| (zsum.north);
      \draw[fl] (ge.east) -| (zsum.south);

      \node[lab, left=14mm of ce] (inX) {$\mathbf{X}_t$};
      \node[lab, left=14mm of le] (inL) {$\boldsymbol{\ell}_t$};
      \node[lab, left=14mm of sc] (inF) {$\mathcal{C}$};
      \node[lab, left=14mm of ge] (inG)
        {$\mathbf{X}_t,\,\boldsymbol{\ell}_t,\,t$};
      \draw[fl] (inX) -- (ce.west);
      \draw[fl] (inL) -- (le.west);
      \draw[fl] (inG) -- (ge.west);
      \coordinate (condbranch) at ([xshift=-7.5mm]sc.west);
      \node[dot] at (condbranch) {};
      \draw[wire] (inF) -- (condbranch);
      \draw[fl] (condbranch) -- (sc.west);
      \draw[fl] (condbranch) |- (pc.west);
      \coordinate (cbr) at ([xshift=3mm]sc.east);
      \node[dot] at (cbr) {};
      \draw[fl] (cbr) |- ([yshift=3.5mm]pc.north) -- (pc.north);
      \coordinate (embtop) at ([yshift=2mm]ce.north);
      \coordinate (embleft) at ([xshift=-9.6mm]ce.west);
      \coordinate (modulebottom) at ([yshift=-10mm]zsum.center);
      \begin{scope}[on background layer]
        \node[modbg, draw=cEmbedDk, fill=cEmbedBg,
              fit=(embleft)(embtop)(modulebottom)(sc)(pc)(ge)(ssum)(zsum)]
              (embbg) {};
      \end{scope}
      \node[mtitle, above=0.7mm of embbg.north] (embtitle) {Input Embedder};

      \node[tri, right=16mm of zsum] (trio) {TriMulOut};
      \node[sum, right=2.6mm of trio] (za) {$+$};
      \node[tri, right=3mm of za] (trii) {TriMulIn};
      \node[sum, right=2.6mm of trii] (zb) {$+$};
      \node[trans, right=3mm of zb] (tz) {Transition};
      \node[sum, right=2.6mm of tz] (zc) {$+$};
      \draw[fl] (zsum) -- (trio.west);
      \draw[fl] (trio) -- (za);
      \draw[fl] (za) -- (trii.west);
      \draw[fl] (trii) -- (zb);
      \draw[fl] (zb) -- (tz.west);
      \draw[fl] (tz) -- (zc);
      \coordinate (rzone) at ([xshift=-4mm]trio.west);
      \coordinate (rztwo) at ([xshift=-1.5mm]trii.west);
      \coordinate (rzthree) at ([xshift=-1.5mm]tz.west);
      \foreach \point in {rzone,rztwo,rzthree} {
        \node[dot] at (\point) {};
      }
      \draw[fl] (rzone) -- ++(0,-\skipdy) -| (za.south);
      \draw[fl] (rztwo) -- ++(0,-\skipdy) -| (zb.south);
      \draw[fl] (rzthree) -- ++(0,-\skipdy) -| (zc.south);

      \node[attn, dashed, anchor=west] (aps) at (trii.west |- ssum) {Attention};
      \node[sum, right=2.6mm of aps] (sa) {$+$};
      \node[trans, dashed, right=3mm of sa] (tps) {Transition};
      \node[sum, right=2.6mm of tps] (sb) {$+$};
      \draw[fl] (ssum) -- (aps.west);
      \draw[fl] (aps) -- (sa);
      \draw[fl] (sa) -- (tps.west);
      \draw[fl] (tps) -- (sb);
      \coordinate (rsone) at ([xshift=-4mm]aps.west);
      \coordinate (rstwo) at ([xshift=-1.5mm]tps.west);
      \node[dot] at (rsone) {};
      \node[dot] at (rstwo) {};
      \draw[fl] (rsone) -- ++(0,\skipdy) -| (sa.north);
      \draw[fl] (rstwo) -- ++(0,\skipdy) -| (sb.north);
      \draw[fl] (zc.north) |- ($(zc.north)!0.5!(aps.south)$) -| (aps.south);
      \coordinate (pmhi) at ([yshift=\skipdy+1.5mm]aps.north);
      \coordinate (pmlo) at ([yshift=-\skipdy-1.5mm]trio.south);
      \coordinate (pairtop) at (trio |- embtop);
      \coordinate (pairbottom) at (trio |- modulebottom);
      \begin{scope}[on background layer]
        \node[modbg, draw=archPairLine, fill=cPairBg,
              fit=(rzone)(trio)(zc)(aps)(sb)(pmhi)(pmlo)(pairtop)(pairbottom)]
              (pmbg) {};
      \end{scope}
      \node[mtitle, above=0.7mm of pmbg.north] (pmtitle)
        {Pairmixer ($\times 4$)};

      \node[lab, right=8mm of sb] (sout) {$\mathbf S$};
      \node[lab, right=8mm of zc] (pout) {$\mathbf P$};
      \draw[fl] (sb) -- (sout.west);
      \draw[fl] (zc) -- (pout.west);
      \coordinate (panelbottom) at ([yshift=-1.6mm]pmbg.south);

      \begin{scope}[on background layer]
        \node[modbg, draw=archFrame, fill=none, inner sep=7pt,
              fit=(embbg)(embtitle)(pmbg)(pmtitle)(inF)(inG)(sout)(pout)
                  (panelbottom)]
              (outerbg) {};
      \end{scope}
    \end{tikzpicture}%

  \end{minipage}\hfill
  \begin{minipage}[t]{204mm}
    \centering
    {\sffamily\bfseries\Large (b) After-Pairmixer noisy-input entry}\\[1.5mm]
    \begin{tikzpicture}[archfig]
      \node[emb] (sc) {SingleCond};
      \node[emb, below=17mm of sc] (pc) {PairCond};
      \node[lab, left=7mm of sc] (inF) {$\mathcal{C}$};
      \coordinate (condbranch) at ([xshift=-3mm]sc.west);
      \node[dot] at (condbranch) {};
      \draw[wire] (inF) -- (condbranch);
      \draw[fl] (condbranch) -- (sc.west);
      \draw[fl] (condbranch) |- (pc.west);
      \coordinate (cbr) at ([xshift=3mm]sc.east);
      \node[dot] at (cbr) {};
      \draw[fl] (cbr) |- ([yshift=3.5mm]pc.north) -- (pc.north);
      \coordinate (embtop) at ([yshift=2mm]sc.north);
      \coordinate (embleft) at ([xshift=-3mm]sc.west);
      \begin{scope}[on background layer]
        \node[modbg, draw=cEmbedDk, fill=cEmbedBg,
              fit=(embleft)(embtop)(sc)(pc)(cbr)] (embbg) {};
      \end{scope}
      \node[mtitle, above=0.7mm of embbg.north] (embtitle) {Input Embedder};

      \node[tri, right=18mm of pc] (trio) {TriMulOut};
      \node[sum, right=2.3mm of trio] (za) {$+$};
      \node[tri, right=2.6mm of za] (trii) {TriMulIn};
      \node[sum, right=2.3mm of trii] (zb) {$+$};
      \node[triattn, right=2.6mm of zb] (tas) {TriAttnStart};
      \node[sum, right=2.3mm of tas] (zc) {$+$};
      \node[triattn, right=2.6mm of zc] (tae) {TriAttnEnd};
      \node[sum, right=2.3mm of tae] (zd) {$+$};
      \node[trans, right=2.6mm of zd] (tz) {Transition};
      \node[sum, right=2.3mm of tz] (ze) {$+$};
      \draw[fl] (pc.east) -- (trio.west);
      \draw[fl] (trio) -- (za);
      \draw[fl] (za) -- (trii.west);
      \draw[fl] (trii) -- (zb);
      \draw[fl] (zb) -- (tas.west);
      \draw[fl] (tas) -- (zc);
      \draw[fl] (zc) -- (tae.west);
      \draw[fl] (tae) -- (zd);
      \draw[fl] (zd) -- (tz.west);
      \draw[fl] (tz) -- (ze);
      \coordinate (rzone) at ([xshift=-4mm]trio.west);
      \coordinate (rztwo) at ([xshift=-1.5mm]trii.west);
      \coordinate (rzthree) at ([xshift=-1.5mm]tas.west);
      \coordinate (rzfour) at ([xshift=-1.5mm]tae.west);
      \coordinate (rzfive) at ([xshift=-1.5mm]tz.west);
      \foreach \point in {rzone,rztwo,rzthree,rzfour,rzfive} {
        \node[dot] at (\point) {};
      }
      \draw[fl] (rzone) -- ++(0,-\skipdy) -| (za.south);
      \draw[fl] (rztwo) -- ++(0,-\skipdy) -| (zb.south);
      \draw[fl] (rzthree) -- ++(0,-\skipdy) -| (zc.south);
      \draw[fl] (rzfour) -- ++(0,-\skipdy) -| (zd.south);
      \draw[fl] (rzfive) -- ++(0,-\skipdy) -| (ze.south);

      \node[attn, dashed, anchor=west] (aps) at (tae.west |- sc) {Attention};
      \node[sum, right=2.6mm of aps] (sa) {$+$};
      \node[trans, dashed, right=3mm of sa] (tps) {Transition};
      \node[sum, right=2.6mm of tps] (sb) {$+$};
      \draw[fl] (sc.east) -- (aps.west);
      \draw[fl] (aps) -- (sa);
      \draw[fl] (sa) -- (tps.west);
      \draw[fl] (tps) -- (sb);
      \coordinate (rsone) at ([xshift=-4mm]aps.west);
      \coordinate (rstwo) at ([xshift=-1.5mm]tps.west);
      \node[dot] at (rsone) {};
      \node[dot] at (rstwo) {};
      \draw[fl] (rsone) -- ++(0,\skipdy) -| (sa.north);
      \draw[fl] (rstwo) -- ++(0,\skipdy) -| (sb.north);
      \draw[fl] (ze.north) |- ($(ze.north)!0.5!(aps.south)$) -| (aps.south);
      \coordinate (pmhi) at ([yshift=\skipdy+1.5mm]aps.north);
      \coordinate (pmlo) at ([yshift=-\skipdy-1.5mm]trio.south);
      \begin{scope}[on background layer]
        \node[modbg, draw=archPairLine, fill=cPairBg,
              fit=(rzone)(trio)(ze)(aps)(sb)(pmhi)(pmlo)] (pmbg) {};
      \end{scope}
      \node[mtitle, above=0.7mm of pmbg.north] (pmtitle)
        {Pairmixer ($\times 4$)};

      \node[sum, right=18mm of ze] (zadd) {$+$};
      \node[sum] (sadd) at (zadd |- sc) {$+$};
      \node[emb, above=3mm of sadd] (ce) {Coord};
      \node[emb, below=3mm of sadd] (le) {Lattice};
      \node[emb, dashed, below=3mm of zadd] (ge) {GEM};
      \draw[fl] (sb) -- (sadd.west);
      \draw[fl] (ce.south) -- (sadd.north);
      \draw[fl] (le.north) -- (sadd.south);
      \draw[fl] (ze) -- (zadd.west);
      \draw[fl] (ge.north) -- (zadd.south);
      \node[lab, left=4mm of ce] (inX) {$\mathbf{X}_t$};
      \node[lab, left=4mm of le] (inL) {$\boldsymbol{\ell}_t$};
      \node[lab, below=5mm of ge] (inG)
        {$\mathbf{X}_t,\,\boldsymbol{\ell}_t,\,t$};
      \draw[fl] (inX) -- (ce.west);
      \draw[fl] (inL) -- (le.west);
      \draw[fl] (inG.north) -- (ge.south);
      \begin{scope}[on background layer]
        \node[modbg, draw=cEmbedDk, fill=cEmbedBg,
              fit=(ce)(le)(ge)(sadd)(zadd)] (noisybg) {};
      \end{scope}
      \node[mtitle, above=0.7mm of noisybg.north] (noisytitle)
        {Noisy Input Embedder};

      \node[lab, right=13mm of sadd] (sout) {$\mathbf S$};
      \node[lab, right=13mm of zadd] (pout) {$\mathbf P$};
      \draw[fl] (sadd) -- (sout.west);
      \draw[fl] (zadd) -- (pout.west);

      \coordinate (cachebraceleft) at (embbg.west |- pmbg.south);
      \coordinate (cachebraceright) at (pmbg.east |- pmbg.south);
      \draw[draw=archFrame!55, line width=0.65pt, densely dotted, decorate,
            decoration={brace, mirror, amplitude=4pt}]
        ([yshift=-3mm]cachebraceleft) -- ([yshift=-3mm]cachebraceright)
        node[midway, below=2mm, font=\scriptsize\itshape, text=archFrame]
        (cachelabel) {Cacheable};

      \begin{scope}[on background layer]
        \node[modbg, draw=archFrame, fill=none, inner sep=7pt,
              fit=(embbg)(embtitle)(pmbg)(pmtitle)(noisybg)(noisytitle)
                  (inF)(inG)(sout)(pout)(cachelabel)] (outerbg) {};
      \end{scope}
    \end{tikzpicture}%
  \end{minipage}
  \end{minipage}
  }
    \caption{\textbf{Model variants for architecture ablations.}
    \textbf{(a)} In the pre-entry variant, noisy coordinate and lattice embeddings
    and optional geometric enhancement module (GEM) features are injected before
    \textsc{Pairmixer}. \textbf{(b)} In the post-entry variant, the condition-only
    \textsc{Input Embedder} and \textsc{Pairmixer} are evaluated before noisy-state
    injection, allowing their outputs to be cached across denoising steps. Modules
    with dotted outlines denote optional components. Both variants produce single
    and pair representations $\mathbf S$ and $\mathbf P$, which are passed to the
    same DiT trunk.}
  \label{fig:sp-architecture}
\end{figure}

\begin{figure}[!t]
  \centering
  \includegraphics[width=\textwidth]{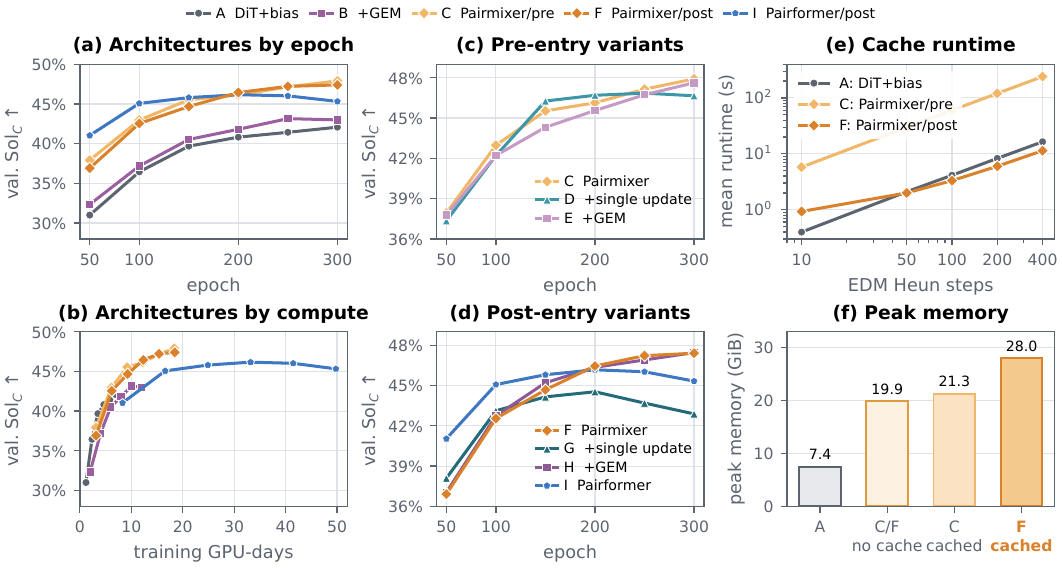}
  \caption{\textbf{Architecture selection, extensions, and caching.}
    Panels~(a)--(d) report validation $\mathrm{Sol}_C$ on the $\le$300-atom
    subset with a generation budget of 30 candidates under EDM Heun 200. Pairmixer gives the
    dominant gain; post-entry Config.~(F) remains close to Config.~(C), while
    single updates and GEM do not improve final coverage. Panel~(e) reports
    runtime for synthetic $[64,300]$ EDM Heun inputs, including
    cache construction, and panel~(f) reports peak memory. Caching post-entry
    condition-only pair states removes the per-step Pairmixer cost at a fixed
    memory increase.}
  \label{fig:architecture-ablation-convergence}
\end{figure}

\textbf{Results and selection.}
Pairmixer outperforms Pairformer and DiT with pair-bias attention
($0.479$ vs.\ $0.462$ vs.\ $0.421$; \Cref{fig:architecture-ablation-convergence}a--b).
Pairformer performs better early in training but later degrades and is
substantially more expensive than Pairmixer. GEM improves DiT with pair-bias
attention, whereas neither GEM nor the single-track update improves Pairmixer
(\Cref{fig:architecture-ablation-convergence}c--d). Pre-entry Pairmixer slightly
outperforms post-entry Pairmixer ($0.479$ vs.\ $0.474$), but caching the post-entry design yields a $20.1\times$ reduction in 200-step sampling time in a synthetic $[64,300]$ benchmark, with peak memory increasing slightly from $21.3$ to $28.0$ GiB (\Cref{fig:architecture-ablation-convergence}e--f). We therefore
select post-entry Pairmixer without additional modules. We report parameter count and GFLOPs of each architecture variant in \Cref{app:detailed-ablation-tables}.

\subsection{Training: Selecting the Flow-Matching Recipe}
\label{sec:recipe-training}

\textbf{Explored designs.}
We ablate each training choice separately. For time sampling, we compare uniform,
logit-normal, Beta, and the Beta--uniform mixture proposed by
\citet{geffner2025proteina}. We also test
random translations in fractional coordinates, wrapping each molecule into the
unit cell by its centroid~\citep{gruver2024finetuned}. We compare $L^1$ and $L^2$ losses and vary the coordinate-to-lattice weighting in Equation~\eqref{eq:method-vfm-loss}.

We further test auxiliary supervision on periodic pair distances. Let
$\mathcal{P}=\{(i,j):i\ne j,\ d_{ij}<15\,\text{\AA}\}$, where $d_{ij}$ and
$\hat d_{ij}$ denote target and predicted minimum-image distances. We consider an
$L^1$ loss following CLARI~\citep{lo2026clari} and a periodic adaptation of
AlphaFold~3 smooth-LDDT~\citep{Abramson2024}:
\begin{equation}
    \mathcal{L}_{\mathrm{pair}}^{L^1}
    = \frac{1}{|\mathcal{P}|}\sum_{(i,j)\in\mathcal{P}}
      \left|\hat d_{ij}-d_{ij}\right|,
    \qquad
    \mathcal{L}_{\mathrm{pair}}^{\mathrm{smooth}}
    = 1-\frac{1}{4|\mathcal{P}|}\sum_{(i,j)\in\mathcal{P}}
      \sum_{\tau\in\{0.5,1,2,4\}}
      \sigma\!\left(\tau-\left|\hat d_{ij}-d_{ij}\right|\right),
    \label{eq:training-pair-losses}
\end{equation}
where $\sigma$ is the sigmoid function and distances are in angstroms.
Finally, we compare AdamW~\citep{loshchilov2019decoupled} with
Muon~\citep{liu2025muon} and evaluate raw versus EMA parameters.

\begin{figure}[!t]
  \centering
  \includegraphics[width=\textwidth]{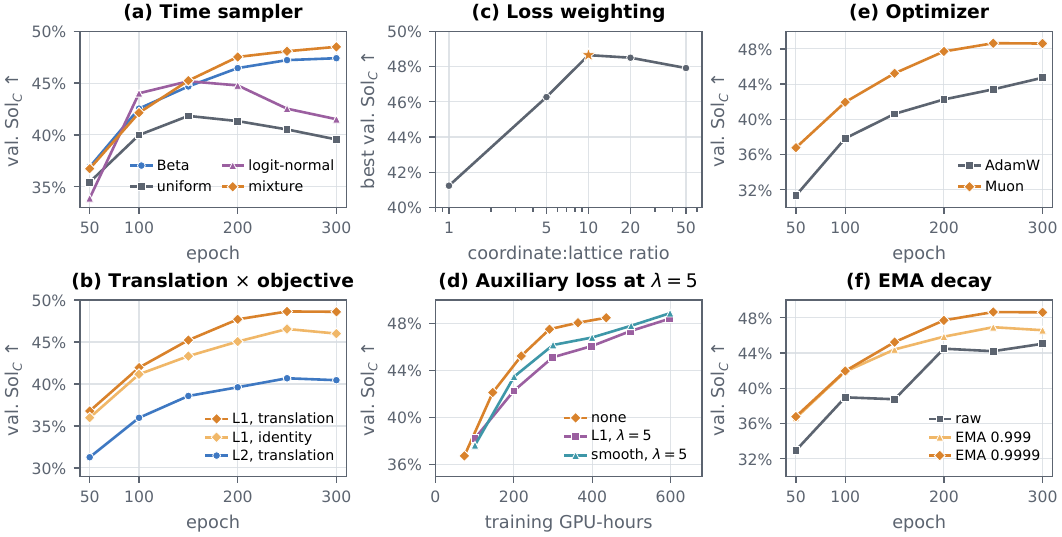}
  \caption{\textbf{Training recipe selection.} All panels report
    validation $\mathrm{Sol}_C$ under EDM Heun 200 with a generation budget of
    30 candidates.
    \textbf{(a)} The Beta--uniform mixture gives the strongest completed score.
    \textbf{(b)} With L1 regression, crystal translation improves the best score
    from $0.466$ to $0.486$; under translation, L1 outperforms L2
    ($0.486$ vs.\ $0.407$). \textbf{(c)} A $10{:}1$ coordinate-to-lattice
    ratio is optimal. \textbf{(d)} At $\lambda{=}5$, smooth-LDDT improves the
    best score by only $0.004$ while increasing training cost from
    436 to 599 GPU-hours. \textbf{(e)} Muon outperforms
    AdamW. \textbf{(f)} EMA decay $0.9999$ outperforms raw weights and decay
    $0.999$.}
  \label{fig:training-recipe-summary}
\end{figure}

\textbf{Results and selection.}
The Beta--uniform mixture outperforms pure Beta, logit-normal, and uniform time
distributions (\Cref{fig:training-recipe-summary}a). Random translation
augmentation improves $\mathrm{Sol}_C$ from $0.466$ to $0.486$, while $L^1$
substantially outperforms $L^2$ ($0.486$ vs.\ $0.407$;
\Cref{fig:training-recipe-summary}b). Increasing the coordinate-to-lattice
weight ratio improves performance from $0.412$ at $1{:}1$ to $0.486$ at
$10{:}1$, after which performance declines
(\Cref{fig:training-recipe-summary}c). Auxiliary pair supervision provides
little benefit: the best smooth-LDDT setting improves $\mathrm{Sol}_C$ by only
$0.004$ while requiring $1.37\times$ more training GPU-hours (599 vs.\ 436). Muon outperforms AdamW ($0.486$ vs.\ $0.447$), and an EMA decay of $0.9999$ outperforms both raw
weights and a decay of $0.999$ (\Cref{fig:training-recipe-summary}e--f). We
therefore select Beta--uniform time sampling, random translation augmentation,
$L^1$ endpoint regression with a $10{:}1$ coordinate-to-lattice weight ratio,
no auxiliary pair loss, Muon, and an EMA decay of $0.9999$.

\subsection{Conditioning: Robustness Through Template Dropout}
\label{sec:recipe-conditioning}

\textbf{Explored designs.}
Because available conditioning information varies across targets, a single model should handle different subsets of optional inputs. We vary
the template-coordinate dropout probability
$p_{\mathrm{tpl}}\in\{0,0.5,1\}$ and evaluate each model under five
inference-time conditioning settings: none, stereochemistry only, template only,
stereochemistry plus template, and all three conditions including space group.

\begin{figure}[!t]
  \centering
  \includegraphics[width=\textwidth]{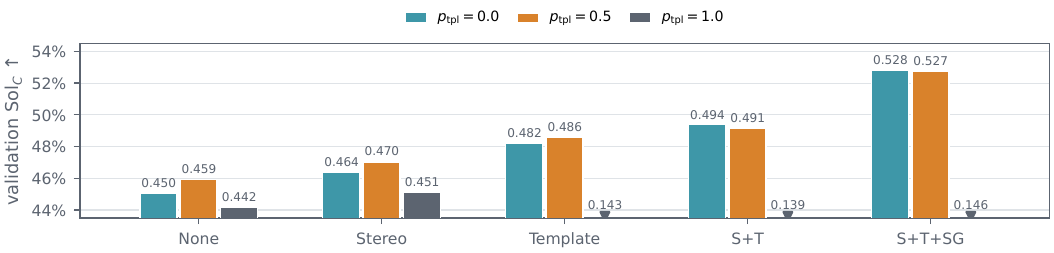}
  \caption{\textbf{Moderate template dropout is most robust across conditioning
    settings.} Validation $\mathrm{Sol}_C$ at epoch 300 using EDM--Heun with
    200 steps and 30 candidates per target. S, T, and SG denote stereochemistry,
    template coordinates, and space-group conditioning, respectively. Among
    models trained with $p_{\mathrm{tpl}}\in\{0,0.5,1\}$,
    $p_{\mathrm{tpl}}{=}0.5$ achieves the highest average performance across the
    five inference-time conditioning settings and is therefore selected.
    Template-conditioned results for $p_{\mathrm{tpl}}{=}1$ fall below the
    plotted range and are shown as downward triangles.}
  \label{fig:conditioning-sweep}
\end{figure}

\textbf{Results and selection.}
Moderate template dropout ($p_{\mathrm{tpl}}=0.5$) is the most robust across
conditioning settings, achieving the highest mean $\mathrm{Sol}_C$ over the five
settings ($0.487$, compared with $0.483$ and $0.264$ for
$p_{\mathrm{tpl}}=0$ and $1$, respectively). Notably, exposure to templates
during training also improves performance when no optional conditioning is
provided: $p_{\mathrm{tpl}}=0.5$ reaches $0.459$, compared with $0.442$ for a
model that never observes templates. We therefore select
$p_{\mathrm{tpl}}=0.5$ (\Cref{fig:conditioning-sweep}).

\subsection{Inference: Efficient Sampling with EDM--Heun}
\label{sec:recipe-inference}

\textbf{Explored designs.}
We compare flow ODE with Euler integration, flow SDE with Euler--Maruyama, and
EDM--Heun~\citep{karras2022elucidating} at matched numbers of function
evaluations (NFE), and sweep the EDM discretization exponent $\rho$. For
template conditioning, we compare random rotation and translation, rotation
only, and no augmentation at inference to test whether reducing template
randomness improves performance. Finally, we evaluate
autoguidance~\citep{karras2024guiding}, using earlier training checkpoints as
degraded models following \citet{geffner2025proteina}.

\textbf{Results and selection.}
At 400 NFE, EDM--Heun achieves $\mathrm{Sol}_C=0.482$, compared with $0.430$
for flow SDE and $0.408$ for flow ODE. Increasing EDM--Heun from 400 to 1000
NFE improves $\mathrm{Sol}_C$ by only $0.005$, so we retain 400 NFE.
Performance is similar for $\rho\in[5,10]$; following
\citet{karras2022elucidating}, we select $\rho=7$. Reducing template
augmentation does not improve performance, with all three variants reaching
$0.485$--$0.487$, so we retain the training-matched rotation and translation.
Autoguidance also provides no benefit: the best guided setting reaches $0.470$,
compared with $0.487$ without guidance. We therefore select EDM--Heun at 400 NFE with
$\rho=7$, template rotation and translation, and no autoguidance.

\subsection{Scaling: Balancing Single- and Pair-Track Widths}
\label{sec:recipe-model-scaling}

\textbf{Explored designs.}
We study how the single-track width $d_s$ and pair-track width $d_p$ affect
performance when scaled independently or together. All other settings, including
depth, Pairmixer placement, conditioning, validation split, and sampler, are held
fixed. We report both the single-sample solve rate $\mathrm{Sol}_S$ and the
30-sample crystal-level solve rate $\mathrm{Sol}_C$, capturing per-sample success
and coverage across targets, respectively.

Starting from the base configuration W0 with $(d_s,d_p)=(512,128)$, we consider:
\begin{itemize}
    \item Pair scaling (W1, W2): increase $d_p$ to $192$ and $256$,
    respectively, while keeping $d_s=512$.
    \item Single scaling (W3, W5): increase $d_s$ to $768$ and $1024$,
    respectively, while keeping $d_p=128$.
    \item Balanced scaling (W4): increase both widths to
    $(d_s,d_p)=(768,192)$, preserving $d_s/d_p=4$.
\end{itemize}

\begin{figure}[t]
  \centering
  \includegraphics[width=\textwidth]{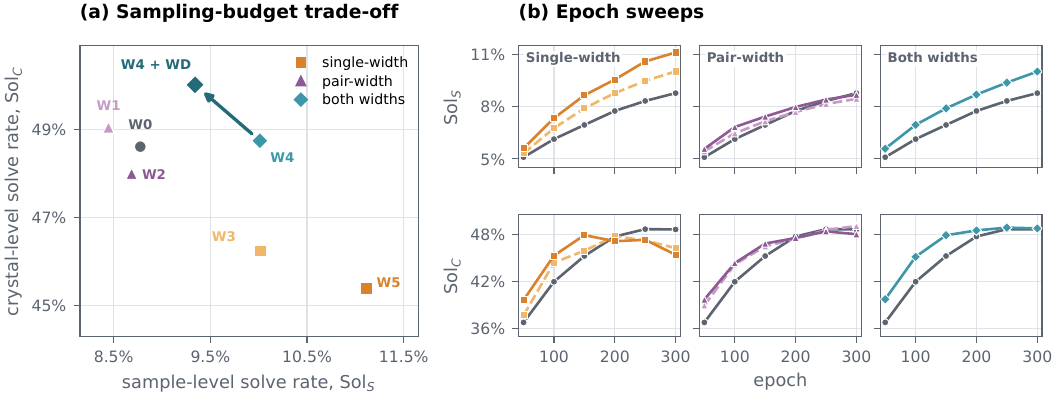}
  \caption{\textbf{Balanced width scaling reaches the sampling-budget Pareto
    frontier.} \textbf{(a)} Single-sample solve rate
    $\mathrm{Sol}_S$ ($k{=}1$) versus crystal-level solve rate
    $\mathrm{Sol}_C$ at $k{=}30$ at epoch 300. Marker shape and color indicate the
    width-scaling family. Scaling either the single or pair width alone does not
    consistently improve the trade-off between per-sample success and target
    coverage, whereas jointly scaling both widths reaches the Pareto frontier.
    Adding weight decay of $10^{-2}$ further shifts the balanced model toward
    higher $k{=}30$ coverage. W0 is selected as \model{}-M, and W4 with weight
    decay (W4 + WD) as \model{}-L. \textbf{(b)} Epoch sweeps for single-width
    scaling (W0, W3, W5), pair-width scaling (W0, W1, W2), and balanced scaling
    (W0, W4). Top and bottom rows report $\mathrm{Sol}_S$ and $\mathrm{Sol}_C$,
    respectively. All metrics use the shared validation protocol with 30
    candidates per target and 200-step EDM--Heun sampling.}
  \label{fig:width-scaling-tradeoff}
\end{figure}

\textbf{Scaling either width alone is insufficient.}
The epoch sweeps reveal distinct limitations of the two isolated scaling paths.
Increasing only the single width improves $\mathrm{Sol}_S$ but eventually reduces
$\mathrm{Sol}_C$, suggesting that the added capacity makes successes more
repeatable on already-solvable targets rather than expanding coverage. Increasing
only the pair width, in contrast, yields no consistent improvement in either
metric. Thus, neither isolated scaling path provides a uniformly better trade-off
across sampling budgets
(\Cref{fig:width-scaling-tradeoff}; \Cref{tab:width_scaling_ablation}).

\textbf{Balanced scaling reaches the sampling-budget Pareto frontier.}
Jointly scaling both widths allows W4 to maintain single-sample success while
recovering crystal-level coverage, yielding a better trade-off than scaling either
width alone. Adding weight decay of $10^{-2}$ shifts W4 toward greater coverage,
trading some success at $k{=}1$ for improved coverage at $k{=}30$. We prioritize
coverage because CSP workflows generate multiple candidates for subsequent ranking;
a structure absent from the candidate set cannot be recovered downstream
\citep{hunnisett2024seventhranking,hunnisett2024seventhgeneration}.
We therefore retain W0 as \model{}-M and select W4 with weight decay
$10^{-2}$ as \model{}-L (\Cref{fig:width-scaling-tradeoff}).

\begin{table}[!t]
\centering
\caption{\textbf{Width-scaling ablation.}
All rows keep depth, Pairmixer placement, conditioning, validation split, and
sampler fixed while varying the single width $d_s$ and pair width $d_p$.
$\mathrm{Sol}_S$ and $\mathrm{Sol}_C$ are evaluated at epoch 300 with 30
candidates per target and 200-step EDM--Heun sampling (400 NFE). GFLOPs denote
analytical denoiser-forward costs at $N{=}300$. Teal rows mark the selected width
configurations: W0 defines \model{}-M, while W4 defines the width of \model{}-L
before applying weight decay as selected in \Cref{fig:width-scaling-tradeoff}.}
\label{tab:width_scaling_ablation}
\small
\begin{tabular}{@{}lccccccc@{}}
\toprule
ID & $d_s$ & $d_p$ & Heads & Params (M) & GFLOPs &
$\mathrm{Sol}_S \uparrow$ & $\mathrm{Sol}_C \uparrow$ \\
\midrule
\rowcolor{mfTealFill}
W0 & 512 & 128 & 8 & 87.56 & 379.6 & 0.088 & 0.486 \\
W1 & 512 & 192 & 8 & 89.65 & 762.6 & 0.084 & 0.490 \\
W2 & 512 & 256 & 8 & 92.54 & 1287.2 & 0.087 & 0.480 \\
W3 & 768 & 128 & 12 & 184.79 & 421.0 & 0.100 & 0.462 \\
\rowcolor{mfTealFill}
W4 & 768 & 192 & 12 & 186.91 & 804.7 & 0.100 & 0.487 \\
W5 & 1024 & 128 & 16 & 330.79 & 483.8 & 0.111 & 0.454 \\
\bottomrule
\end{tabular}
\end{table}

\section{Conclusion}

We present \model{}, a flexible all-atom generator for molecular CSP based on
variational flow matching. \model{} supports multi-component and organometallic
crystals and optional conditioning on molecular templates, stereochemistry, and
space-group information. Controlled studies of architecture, training,
conditioning, inference, and scaling yield a practical recipe based on cacheable
Pairmixer representations, $L^1$ endpoint regression, EDM--Heun sampling, and
balanced single- and pair-track capacity. Inspired by the CCDC CSP blind tests, we introduce a two-track evaluation that separates generator-only coverage from end-to-end performance under a common downstream relaxation and ranking pipeline. Across six benchmarks, \model{} achieves the best matched-budget
coverage, while under the shared downstream pipeline it achieves higher
experimental-form recovery, lower ranks, and faster convergence than the baselines.

\textbf{Limitations and discussion.}
First, our controlled ablations vary design choices one at a time, 
and therefore do not capture interactions among architecture, training,
conditioning, inference, and scaling; the selected recipe is not guaranteed to be
globally optimal. Second, our ablations and model selection primarily use
$k{=}30$ candidates, although the best model may depend on the candidate budget.
Our scaling results show a trade-off between small-budget success and larger-budget
coverage, suggesting that model selection may benefit from larger candidate pools.
Determining an appropriate selection budget therefore remains an important
question for generative CSP.

\section*{Acknowledgements}
This research was supported by the NVIDIA Academic Grant Program and by the
Advanced GPU Utilization Support Program, funded by the Government of the
Republic of Korea through the Ministry of Science and ICT. This work was also
supported by the Basic Science Research Program through the National Research
Foundation of Korea (NRF), funded by the Ministry of Education
(RS-2025-25435147); grants from the Institute for Information \& Communications
Technology Planning \& Evaluation (IITP), funded by the Korean government
(MSIT), including the Artificial Intelligence Graduate School Program (KAIST;
RS-2019-II190075) and the AI Star Fellowship (KAIST; RS-2025-02304967); a grant
from the Korea Health Industry Development Institute (KHIDI), funded by the
Ministry of Health \& Welfare (MOHW), for Developing a Highly Multimodal (Drug,
Protein, Gene, Cell Imaging, Literature) Foundation Model for ADMET Property
Prediction (No. N0425208); and an NRF grant funded by MSIT
(No. RS-2022-NR072184). We also acknowledge the use of OpenAI's ChatGPT and Codex~\citep{openai_chatgpt,openai_codex} for assistance with research code development, figure preparation, and manuscript revision.


\clearpage
\bibliographystyle{abbrvnat}
\bibliography{ref}

@article{albergo2025stochastic,
  title   = {Stochastic Interpolants: A Unifying Framework for Flows and Diffusions},
  author  = {Albergo, Michael S. and Boffi, Nicholas M. and Vanden-Eijnden, Eric},
  journal = {Journal of Machine Learning Research},
  volume  = {26},
  number  = {209},
  pages   = {1--80},
  year    = {2025}
}

@article{wang2020improving,
  title={Improving conformer generation for small rings and macrocycles based on distance geometry and experimental torsional-angle preferences},
  author={Wang, Shuzhe and Witek, Jagna and Landrum, Gregory A and Riniker, Sereina},
  journal={Journal of chemical information and modeling},
  volume={60},
  number={4},
  pages={2044--2058},
  year={2020},
  publisher={ACS Publications}
}

@article{krivy1976unified,
  title   = {A unified algorithm for determining the reduced ({Niggli}) cell},
  author  = {K{\v{r}}iv{\'y}, I. and Gruber, B.},
  journal = {Acta Crystallographica Section A},
  volume  = {32},
  number  = {2},
  pages   = {297--298},
  year    = {1976},
  doi     = {10.1107/S0567739476000636}
}

@misc{liu2025muon,
  title        = {Muon is Scalable for LLM Training},
  author       = {Jingyuan Liu and Jianlin Su and Xingcheng Yao and Zhejun Jiang
                  and Guokun Lai and Yulun Du and Yidao Qin and Weixin Xu
                  and Enzhe Lu and Junjie Yan and Yanru Chen and Huabin Zheng
                  and Yibo Liu and Shaowei Liu and Bohong Yin and Weiran He
                  and Han Zhu and Yuzhi Wang and Jianzhou Wang and Mengnan Dong
                  and Zheng Zhang and Yongsheng Kang and Hao Zhang and Xinran Xu
                  and Yutao Zhang and Yuxin Wu and Xinyu Zhou and Zhilin Yang},
  year         = {2025},
  eprint       = {2502.16982},
  archivePrefix= {arXiv},
  primaryClass = {cs.LG},
  doi          = {10.48550/arXiv.2502.16982}
}

@inproceedings{loshchilov2019decoupled,
  title     = {Decoupled Weight Decay Regularization},
  author    = {Loshchilov, Ilya and Hutter, Frank},
  booktitle = {International Conference on Learning Representations},
  year      = {2019}
}

@inproceedings{tancik2020fourier,
  title     = {Fourier Features Let Networks Learn High Frequency Functions in Low Dimensional Domains},
  author    = {Tancik, Matthew and Srinivasan, Pratul P. and Mildenhall, Ben and Fridovich-Keil, Sara and Raghavan, Nithin and Singhal, Utkarsh and Ramamoorthi, Ravi and Barron, Jonathan T. and Ng, Ren},
  booktitle = {Advances in Neural Information Processing Systems},
  volume    = {33},
  year      = {2020}
}

@inproceedings{wood2025family,
  title={UMA: A Family of Universal Models for Atoms},
  author={Wood, Brandon M and Dzamba, Misko and Fu, Xiang and Gao, Meng and Shuaibi, Muhammed and Barroso-Luque, Luis and Abdelmaqsoud, Kareem and Gharakhanyan, Vahe and Kitchin, John R and Levine, Daniel S and others},
  booktitle={Advances in Neural Information Processing Systems},
  year={2025},
  url={https://openreview.net/forum?id=SvopaNxYWt}
}

@inproceedings{kim2025flexible,
  title={Flexible {MOF} Generation with Torsion-Aware Flow Matching},
  author={Kim, Nayoung and Kim, Seongsu and Ahn, Sungsoo},
  booktitle={Advances in Neural Information Processing Systems},
  year={2025},
  url={https://openreview.net/forum?id=cLJfumTWLI}
}

@inproceedings{jiao2025mof,
  title={{MOF-BFN}: Metal-organic frameworks structure prediction via bayesian flow networks},
  author={Jiao, Rui and Wu, Hanlin and Huang, Wenbing and Song, Yuxuan and Ouyang, Yawen and Rong, Yu and Xu, Tingyang and Wang, Pengju and Zhou, Hao and Ma, Wei-Ying and others},
  booktitle={The Thirty-ninth Annual Conference on Neural Information Processing Systems},
  year={2025}
}

@inproceedings{wu2025periodic,
  title={A Periodic Bayesian Flow for Material Generation},
  author={Wu, Hanlin and Song, Yuxuan and Gong, Jingjing and Cao, Ziyao and Ouyang, Yawen and Zhang, Jianbing and Zhou, Hao and Ma, Wei-Ying and Liu, Jingjing},
  booktitle={The Thirteenth International Conference on Learning Representations},
  year={2025}
}

@inproceedings{hollmeropen,
  title={Open Materials Generation with Stochastic Interpolants},
  author={H{\"o}llmer, Philipp and Egg, Thomas and Martirossyan, Maya and Fuemmeler, Eric and Shui, Zeren and Gupta, Amit and Prakash, Pawan and Roitberg, Adrian and Liu, Mingjie and Karypis, George and others},
  booktitle={Forty-second International Conference on Machine Learning},
  year={2025}
}

@article{eijkelboom2024variational,
  title={Variational flow matching for graph generation},
  author={Eijkelboom, Floor and Bartosh, Grigory and Andersson Naesseth, Christian and Welling, Max and van de Meent, Jan-Willem},
  journal={Advances in Neural Information Processing Systems},
  volume={37},
  pages={11735--11764},
  year={2024}
}

@article{jiao2024crystal,
  title={Crystal structure prediction by joint equivariant diffusion},
  author={Jiao, Rui and Huang, Wenbing and Lin, Peijia and Han, Jiaqi and Chen, Pin and Lu, Yutong and Liu, Yang},
  journal={Advances in Neural Information Processing Systems},
  volume={36},
  year={2024}
}

@inproceedings{kim2025mofflow,
  title={{MOFFlow}: Flow Matching for Structure Prediction of Metal-Organic Frameworks},
  author={Kim, Nayoung and Kim, Seongsu and Kim, Minsu and Park, Jinkyoo and Ahn, Sungsoo},
  booktitle={The Thirteenth International Conference on Learning Representations},
  year={2025},
  url={https://openreview.net/forum?id=dNT3abOsLo}
}

@inproceedings{geffner2025proteina,
    title={Proteina: Scaling Flow-based Protein Structure Generative Models},
    author={Tomas Geffner and Kieran Didi and Zuobai Zhang and Danny Reidenbach and Zhonglin Cao and Jason Yim and Mario Geiger and Christian Dallago and Emine Kucukbenli and Arash Vahdat and Karsten Kreis},
    booktitle={International Conference on Learning Representations (ICLR)},
    year={2025}
}

@inproceedings{karras2022elucidating,
  title={Elucidating the Design Space of Diffusion-Based Generative Models},
  author={Karras, Tero and Aittala, Miika and Aila, Timo and Laine, Samuli},
  booktitle={Advances in Neural Information Processing Systems},
  volume={35},
  pages={26565--26577},
  year={2022}
}

@article{karras2024guiding,
  title={Guiding a Diffusion Model with a Bad Version of Itself},
  author={Karras, Tero and Aittala, Miika and Kynkaanniemi, Tuomas and Lehtinen, Jaakko and Aila, Timo and Laine, Samuli},
  journal={arXiv preprint arXiv:2406.02507},
  year={2024}
}

@inproceedings{gruver2024finetuned,
  title={Fine-Tuned Language Models Generate Stable Inorganic Materials as Text},
  author={Gruver, Nate and Sriram, Anuroop and Madotto, Andrea and Wilson, Andrew Gordon and Zitnick, C. Lawrence and Ulissi, Zachary},
  booktitle={International Conference on Learning Representations},
  year={2024},
}

@misc{rdkit,
  title={RDKit: Open-source cheminformatics},
  author={Landrum, Greg and others},
  year={2006},
  publisher={Zenodo}
}

@article{Abramson2024,
  author  = {Abramson, Josh and Adler, Jonas and Dunger, Jack and Evans, Richard and Green, Tim and Pritzel, Alexander and Ronneberger, Olaf and Willmore, Lindsay and Ballard, Andrew J. and Bambrick, Joshua and Bodenstein, Sebastian W. and Evans, David A. and Hung, Chia-Chun and O’Neill, Michael and Reiman, David and Tunyasuvunakool, Kathryn and Wu, Zachary and Žemgulytė, Akvilė and Arvaniti, Eirini and Beattie, Charles and Bertolli, Ottavia and Bridgland, Alex and Cherepanov, Alexey and Congreve, Miles and Cowen-Rivers, Alexander I. and Cowie, Andrew and Figurnov, Michael and Fuchs, Fabian B. and Gladman, Hannah and Jain, Rishub and Khan, Yousuf A. and Low, Caroline M. R. and Perlin, Kuba and Potapenko, Anna and Savy, Pascal and Singh, Sukhdeep and Stecula, Adrian and Thillaisundaram, Ashok and Tong, Catherine and Yakneen, Sergei and Zhong, Ellen D. and Zielinski, Michal and Žídek, Augustin and Bapst, Victor and Kohli, Pushmeet and Jaderberg, Max and Hassabis, Demis and Jumper, John M.},
  journal = {Nature},
  title   = {Accurate structure prediction of biomolecular interactions with {AlphaFold} 3},
  year    = {2024},
  volume  = {630},
  number  = {8016},
  pages   = {493--500},
  doi     = {10.1038/s41586-024-07487-w}
}

@article{zaghen2025riemannian,
  title={Riemannian Variational Flow Matching for Material and Protein Design},
  author={Zaghen, Olga and Eijkelboom, Floor and Pouplin, Alison and Liu, Cong and Welling, Max and van de Meent, Jan-Willem and Bekkers, Erik J},
  journal={arXiv preprint arXiv:2502.12981},
  year={2025}
}

@inproceedings{peebles2023scalable,
  title={Scalable diffusion models with transformers},
  author={Peebles, William and Xie, Saining},
  booktitle={Proceedings of the IEEE/CVF international conference on computer vision},
  pages={4195--4205},
  year={2023}
}

@article{bytedance2025protenix,
  title={Protenix-advancing structure prediction through a comprehensive {AlphaFold} 3 reproduction},
  author={ByteDance AML AI4Science Team and Chen, Xinshi and Zhang, Yuxuan and Lu, Chan and Ma, Wenzhi and Guan, Jiaqi and Gong, Chengyue and Yang, Jincai and Zhang, Hanyu and Zhang, Ke and others},
  journal={BioRxiv},
  pages={2025--01},
  year={2025},
  publisher={Cold Spring Harbor Laboratory}
}

@article{yi2025seedfold,
  title={SeedFold: Scaling Biomolecular Structure Prediction},
  author={Yi, Zhou and Chan, Lu and Yiming, Ma and Wei, Qu and Fei, Ye and Kexin, Zhang and Lan, Wang and Minrui, Gui and Quanquan, Gu},
  journal={arXiv preprint arXiv:2512.24354},
  year={2025}
}

@article{ong2013pymatgen,
  title = {Python Materials Genomics (pymatgen): A Robust, Open-Source Python Library for Materials Analysis},
  author = {Ong, Shyue Ping and Richards, William Davidson and Jain, Anubhav and Hautier, Geoffroy and Kocher, Michael and Cholia, Shreyas and Gunter, Dan and Chevrier, Vincent and Persson, Kristin A. and Ceder, Gerbrand},
  journal = {Computational Materials Science},
  volume = {68},
  pages = {314-319},
  year = {2013},
  doi = {10.1016/j.commatsci.2012.10.028}
}

@inproceedings{millerflowmm,
  title={FlowMM: Generating Materials with Riemannian Flow Matching},
  author={Miller, Benjamin Kurt and Chen, Ricky TQ and Sriram, Anuroop and Wood, Brandon M},
  booktitle={Forty-first International Conference on Machine Learning},
  year={2024}
}

@article{luo2025crystalflow,
  title={{CrystalFlow}: A flow-based generative model for crystalline materials},
  author={Luo, Xiaoshan and Wang, Zhenyu and Wang, Qingchang and Shao, Xuechen and Lv, Jian and Wang, Lei and Wang, Yanchao and Ma, Yanming},
  journal={Nature Communications},
  volume={16},
  pages={9267},
  year={2025},
  doi={10.1038/s41467-025-64364-4}
}

@inproceedings{
xie2022crystal,
title={Crystal Diffusion Variational Autoencoder for Periodic Material Generation},
author={Tian Xie and Xiang Fu and Octavian-Eugen Ganea and Regina Barzilay and Tommi S. Jaakkola},
booktitle={International Conference on Learning Representations},
year={2022},
url={https://openreview.net/forum?id=03RLpj-tc_}
}

@inproceedings{
    lipman2023flow,
    title={Flow Matching for Generative Modeling},
    author={Yaron Lipman and Ricky T. Q. Chen and Heli Ben-Hamu and Maximilian Nickel and Matthew Le},
    booktitle={The Eleventh International Conference on Learning Representations },
    year={2023},
    url={https://openreview.net/forum?id=PqvMRDCJT9t}
}

@article{grosse2004numerically,
  title={Numerically stable algorithms for the computation of reduced unit cells},
  author={Grosse-Kunstleve, Ralf W and Sauter, Nicholas K and Adams, Paul D},
  journal={Acta Crystallographica Section A: Foundations of Crystallography},
  volume={60},
  number={1},
  pages={1--6},
  year={2004},
  publisher={International Union of Crystallography}
}

@article{larsen2017atomic,
  title={The atomic simulation environment: a Python library for working with atoms},
  author={Larsen, Ask Hjorth and Mortensen, Jens J{\o}rgen and Blomqvist, Jakob and Castelli, Ivano E and Christensen, Rune and Du{\l}ak, Marcin and Friis, Jesper and Groves, Michael N and Hammer, Bj{\o}rk and Hargus, Cory and others},
  journal={Journal of Physics: Condensed Matter},
  volume={29},
  number={27},
  pages={273002},
  year={2017},
  publisher={IOP Publishing}
}

@article{kim2026atommof,
  title={{AtomMOF}: All-Atom Flow Matching for {MOF}-Adsorbate Structure Prediction},
  author={Kim, Nayoung and Kim, Honghui and Yu, Sihyun and Kim, Minkyu and Kim, Seongsu and Ahn, Sungsoo},
  journal={arXiv preprint arXiv:2602.07351},
  year={2026},
  url={https://arxiv.org/abs/2602.07351}
}

@article{day2009significant,
  title={Significant progress in predicting the crystal structures of small organic molecules---a report on the fourth blind test},
  author={Day, Graeme M. and Cooper, Timothy G. and Cruz-Cabeza, Aurora J. and Hejczyk, Katie E. and Ammon, Herman L. and Boerrigter, Stephan X. M. and Tan, Jonathan S. and Della Valle, Raffaele G. and Venuti, Elisabetta and Jose, K. V. J. and others},
  journal={Acta Crystallographica Section B: Structural Science},
  volume={65},
  number={2},
  pages={107--125},
  year={2009},
  doi={10.1107/S0108768109004066}
}

@article{beran2023frontiers,
  title={Frontiers of molecular crystal structure prediction for pharmaceuticals and functional organic materials},
  author={Beran, Gregory J. O.},
  journal={Chemical Science},
  volume={14},
  number={46},
  pages={13290--13312},
  year={2023},
  doi={10.1039/D3SC03903J}
}

@article{reilly2016sixth,
  title={Report on the sixth blind test of organic crystal structure prediction methods},
  author={Reilly, Anthony M. and Cooper, Richard I. and Adjiman, Claire S. and Bhattacharya, Saswata and Boese, A. Daniel and Brandenburg, Jan Gerit and Bygrave, Peter J. and Bylsma, Rita and Campbell, Josh E. and Car, Roberto and others},
  journal={Acta Crystallographica Section B: Structural Science, Crystal Engineering and Materials},
  volume={72},
  number={4},
  pages={439--459},
  year={2016},
  doi={10.1107/S2052520616007447}
}

@article{hunnisett2024seventhgeneration,
  title={The seventh blind test of crystal structure prediction: structure generation methods},
  author={Hunnisett, Lily M. and Nyman, Jonas and Francia, Nicholas and Abraham, Nathan S. and Adjiman, Claire S. and Aitipamula, Srinivasulu and Alkhidir, Tamador and Almehairbi, Mubarak and Anelli, Andrea and Anstine, Dylan M. and others},
  journal={Acta Crystallographica Section B: Structural Science, Crystal Engineering and Materials},
  volume={80},
  number={6},
  pages={517--547},
  year={2024},
  doi={10.1107/S2052520624007492}
}

@article{hunnisett2024seventhranking,
  title={The seventh blind test of crystal structure prediction: structure ranking methods},
  author={Hunnisett, Lily M. and Francia, Nicholas and Nyman, Jonas and Abraham, Nathan S. and Aitipamula, Srinivasulu and Alkhidir, Tamador and Almehairbi, Mubarak and Anelli, Andrea and Anstine, Dylan M. and Anthony, John E. and others},
  journal={Acta Crystallographica Section B: Structural Science, Crystal Engineering and Materials},
  volume={80},
  number={6},
  pages={548--574},
  year={2024},
  doi={10.1107/S2052520624008679}
}

@article{bardwell2011fifth,
  title={Towards crystal structure prediction of complex organic compounds: a report on the fifth blind test},
  author={Bardwell, David A. and Adjiman, Claire S. and Arnautova, Yelena A. and Bartashevich, Ekaterina and Boerrigter, Stephan X. M. and Braun, Doris E. and Cruz-Cabeza, Aurora J. and Day, Graeme M. and Della Valle, Raffaele G. and Desiraju, Gautam R. and others},
  journal={Acta Crystallographica Section B: Structural Science, Crystal Engineering and Materials},
  volume={67},
  number={6},
  pages={535--551},
  year={2011},
  doi={10.1107/S0108768111042868}
}

@article{battle2010applications,
  title={Applications of the {Cambridge Structural Database} in chemical education},
  author={Battle, Gary M. and Ferrence, Gregory M. and Allen, Frank H.},
  journal={Journal of Applied Crystallography},
  volume={43},
  number={5},
  pages={1208--1223},
  year={2010},
  doi={10.1107/S0021889810024155}
}

@article{groom2016cambridge,
  title={{The Cambridge Structural Database}},
  author={Groom, Colin R. and Bruno, Ian J. and Lightfoot, Matthew P. and Ward, Suzanna C.},
  journal={Acta Crystallographica Section B: Structural Science, Crystal Engineering and Materials},
  volume={72},
  number={2},
  pages={171--179},
  year={2016},
  doi={10.1107/S2052520616003954}
}

@article{sykes2024scripting,
  title={What has scripting ever done for us? {The CSD Python} application programming interface ({API})},
  author={Sykes, Richard A. and Johnson, Natalie T. and Kingsbury, Christopher J. and Harter, J{\"u}rgen and Maloney, Andrew G. P. and Sugden, Isaac J. and Ward, Suzanna C. and Bruno, Ian J. and Adcock, Stewart A. and Wood, Peter A. and McCabe, Patrick and Moldovan, Alexandru A. and Atkinson, Francis and Giangreco, Ilenia and Cole, Jason C.},
  journal={Journal of Applied Crystallography},
  volume={57},
  number={4},
  pages={1235--1250},
  year={2024},
  doi={10.1107/S1600576724005934}
}

@article{chisholm2005compack,
  title={{COMPACK}: a program for identifying crystal structure similarity using distances},
  author={Chisholm, James Alexander and Motherwell, Sam},
  journal={Journal of Applied Crystallography},
  volume={38},
  number={1},
  pages={228--231},
  year={2005},
  doi={10.1107/S0021889804027074}
}

@article{rego2015_3dmol,
  title={{3Dmol.js}: molecular visualization with {WebGL}},
  author={Rego, Nicholas and Koes, David},
  journal={Bioinformatics},
  volume={31},
  number={8},
  pages={1322--1324},
  year={2015},
  doi={10.1093/bioinformatics/btu829}
}

@article{price2014predicting,
  title={Predicting crystal structures of organic compounds},
  author={Price, Sarah L.},
  journal={Chemical Society Reviews},
  volume={43},
  pages={2098--2111},
  year={2014},
  doi={10.1039/C3CS60279F}
}

@article{li2018genarris,
  title={Genarris: Random generation of molecular crystal structures and fast screening with a {Harris} approximation},
  author={Li, Xiayue and Curtis, Farren S. and Rose, Timothy and Schober, Christoph and V{\'a}zquez-Mayagoitia, {\'A}lvaro and Reuter, Karsten and Oberhofer, Harald and Marom, Noa},
  journal={The Journal of Chemical Physics},
  volume={148},
  number={24},
  pages={241701},
  year={2018},
  doi={10.1063/1.5014038}
}

@article{curtis2018gator,
  title={{GAtor}: A first-principles genetic algorithm for molecular crystal structure prediction},
  author={Curtis, Farren and Li, Xiayue and Rose, Timothy and V{\'a}zquez-Mayagoitia, {\'A}lvaro and Bhattacharya, Saswata and Ghiringhelli, Luca M. and Marom, Noa},
  journal={Journal of Chemical Theory and Computation},
  volume={14},
  number={4},
  pages={2246--2264},
  year={2018},
  doi={10.1021/acs.jctc.7b01152}
}

@article{jin2025oxtal,
  title={{OXtal}: An All-Atom Diffusion Model for Organic Crystal Structure Prediction},
  author={Jin, Emily and Nica, Andrei Cristian and Galkin, Mikhail and Rector-Brooks, Jarrid and Lee, Kin Long Kelvin and Miret, Santiago and Arnold, Frances H. and Bronstein, Michael and Bose, Avishek Joey and Tong, Alexander and Liu, Cheng-Hao},
  journal={arXiv preprint arXiv:2512.06987},
  year={2025}
}

@article{subramanian2026packflow,
  title={{PackFlow}: Generative Molecular Crystal Structure Prediction via Reinforcement Learning Alignment},
  author={Subramanian, Akshay and Pan, Elton and Nam, Juno and Weiler, Maurice and Qu, Shuhui and Park, Cheol Woo and Jaakkola, Tommi S. and Olivetti, Elsa and Gomez-Bombarelli, Rafael},
  journal={arXiv preprint arXiv:2602.20140},
  year={2026}
}

@article{zeng2026molcrystalflow,
  title={{MolCrystalFlow}: Molecular Crystal Structure Prediction via Flow Matching},
  author={Zeng, Cheng and Sullivan, Harry W. and Egg, Thomas and Martirossyan, Maya M. and H{\"o}llmer, Philipp and Jin, Jirui and Hennig, Richard G. and Roitberg, Adrian and Martiniani, Stefano and Tadmor, Ellad B. and Liu, Mingjie},
  journal={arXiv preprint arXiv:2602.16020},
  year={2026}
}

@article{lo2026clari,
  title={Fast Organic Crystal Structure Prediction with Unit Cell Flow Matching},
  author={Lo, Alston and Mucko, Luka and Cheng, Austin H. and Cai, Andy and Price, Alastair J. A. and Matusik, Wojciech and Aspuru-Guzik, Al{\'a}n},
  journal={arXiv preprint arXiv:2606.03199},
  year={2026}
}

@article{gharakhanyan2025fastcsp,
  title={{FastCSP}: Accelerated Molecular Crystal Structure Prediction with Universal Model for Atoms},
  author={Gharakhanyan, Vahe and Yang, Yi and Barroso-Luque, Luis and Shuaibi, Muhammed and Levine, Daniel S. and Michel, Kyle and others},
  journal={arXiv preprint arXiv:2508.02641},
  year={2025}
}

@article{veljkovic2026crystalite,
  title={{Crystalite}: A Lightweight Transformer for Efficient Crystal Modeling},
  author={Veljkovi{\'c}, Tin Had{\v z}i and Rosenthal, Joshua and Lon{\v c}ari{\'c}, Ivor and van de Meent, Jan-Willem},
  journal={arXiv preprint arXiv:2604.02270},
  year={2026}
}

@article{ouyangzhang2025triangle,
  title={Triangle Multiplication Is All You Need for Biomolecular Structure Representations},
  author={Ouyang-Zhang, Jeffrey and Murugan, Pranav and Diaz, Daniel J. and Scarpellini, Gianluca and Bowen, Richard Strong and Gruver, Nate and Klivans, Adam and Kr{\"a}henb{\"u}hl, Philipp and Faust, Aleksandra and Al-Shedivat, Maruan},
  journal={arXiv preprint arXiv:2510.18870},
  year={2025}
}

@misc{openai_chatgpt,
  title={{ChatGPT}: Optimizing Language Models for Dialogue},
  author={{OpenAI}},
  year={2022},
  howpublished={\url{https://openai.com/index/chatgpt/}},
  note={Accessed: 2026-08-23}
}

@misc{openai_codex,
  title={Introducing {Codex}},
  author={{OpenAI}},
  year={2025},
  howpublished={\url{https://openai.com/index/introducing-codex/}},
  note={Accessed: 2026-08-23}
}


\appendix
\clearpage
\section{Model Architecture Details}
\label{app:model-architecture}

\algnewcommand{\AlgPhase}[1]{\Statex \texttt{\# #1}}

\textbf{Notation.}
This section specifies the \model{} forward pass in algorithmic form. Indices $i,j,k$ denote atoms, and $h$ denotes an attention head. We use $\mathbf S$ and $\mathbf P$ for single and pair representations, respectively. Symbols $E$, $W$, and $\operatorname{MLP}$ denote learned embedding functions, linear maps, and multilayer perceptrons, respectively, while $\operatorname{Norm}$ denotes normalization. $\operatorname{Split}$ and $\operatorname{Concat}$ partition and concatenate feature channels, respectively. We use $\sigma$ for the sigmoid function, $\phi$ for the configured activation function, $\odot$ for elementwise multiplication, and $d_h$ for the dimension of one attention head.

\textbf{Overall forward pass.} \Cref{alg:model-forward} summarizes the complete forward pass: the model embeds the condition, refines pair features, injects the noisy crystal state, applies the DiT trunk, and predicts coordinate and lattice.

\begin{algorithm}[H]
\caption{Model forward pass}
\label{alg:model-forward}
\begin{algorithmic}[1]
\Require Noisy state $(\mathbf X_t,\boldsymbol\ell_t)$, time $t$, condition
$\mathcal C$, and valid-atom mask $\mathbf m$
\Ensure Posterior mean $(\widehat{\mathbf X}_1,\widehat{\boldsymbol\ell}_1)$
\AlgPhase{1. Encode the condition and refine pair features}
\State $(\mathbf S_{\mathcal C},\mathbf P)
    \gets \Call{ConditionEmbed}{\mathcal C,\mathbf m}$
\For{each Pairmixer block}
    \State $\mathbf P\gets\Call{PairmixerBlock}{\mathbf P,\mathbf m}$
\EndFor
\State $\mathbf P^\star\gets\mathbf P$
\Statex
\AlgPhase{2. Inject the noisy state}
\State $(\mathbf S,\mathbf c_t)
    \gets \Call{NoisyStateEmbed}{\mathbf S_{\mathcal C},\mathbf X_t,
    \boldsymbol\ell_t,t}$
\Statex
\AlgPhase{3. Propagate and decode}
\For{each DiT block}
    \State $\mathbf S
        \gets \Call{DiTBlock}{\mathbf S,\mathbf c_t,\mathbf P^\star,\mathbf m}$
\EndFor
\State $(\widehat{\mathbf X}_1,\widehat{\boldsymbol\ell}_1)
    \gets \Call{PredictionHeads}{\mathbf S,\mathbf m}$
\State \Return $(\widehat{\mathbf X}_1,\widehat{\boldsymbol\ell}_1)$
\end{algorithmic}
\end{algorithm}

\textbf{Condition embedding.} For \Cref{alg:condition-embedding}, $\mathbf a_i$ contains the atomic number, periodic-table descriptors, formal charge, and chirality of atom $i$. Template coordinates are denoted by $\widetilde{\mathbf x}_i$, with availability indicator $m_{\mathrm T}$. Molecular membership, bond type, and bond stereochemistry are denoted by $u_i$, $\tau_{ij}$, and $\zeta_{ij}$, respectively; unavailable categorical conditions use learned null categories. The global condition contains the space-group embedding. Bond, stereochemistry, and template features are restricted to atom pairs within the same molecular component, and stereochemistry features are zero for nonbonded pairs.

\begin{algorithm}[H]
\caption{Condition embedding}
\label{alg:condition-embedding}
\begin{algorithmic}[1]
\Require Condition $\mathcal C$ and valid-atom mask $\mathbf m$
\Ensure Condition-only single representation $\mathbf S_{\mathcal C}$ and
ordered pair representation $\mathbf P$
\AlgPhase{1. Embed atom-level and global conditions}
\State $\displaystyle
\mathbf S_{\mathcal C,i}\gets
E_{\mathrm{atom}}(\mathbf a_i)
+m_{\mathrm T}E_{\mathrm{temp}}(\widetilde{\mathbf x}_i)
+m_iE_{\mathrm{global}}(\mathcal C)$
\Statex
\AlgPhase{2. Construct intramolecular pair features}
\State $M^{\mathrm{mol}}_{ij}\gets m_i m_j\mathbf 1[u_i=u_j]$
\State $\boldsymbol\delta_{ij}
    \gets\widetilde{\mathbf x}_i-\widetilde{\mathbf x}_j$,
    $\rho_{ij}\gets(1+\lVert\boldsymbol\delta_{ij}\rVert_2^2)^{-1}$
\Statex $\displaystyle \overline{\mathbf P}_{ij}\gets W_Q\mathbf S_{\mathcal C,i}+W_K\mathbf S_{\mathcal C,j}+M^{\mathrm{mol}}_{ij}\bigl[E_{\mathrm{bond}}(\tau_{ij})+E_{\mathrm{stereo}}(\zeta_{ij})+m_{\mathrm T}(W_\delta\boldsymbol\delta_{ij}+W_\rho\rho_{ij})\bigr].$
\Statex
\AlgPhase{3. Refine each pair independently}
\State $\mathbf P_{ij}\gets\overline{\mathbf P}_{ij}
    +\operatorname{MLP}_{P}\!\left(
    \operatorname{Norm}_{P}(\overline{\mathbf P}_{ij})\right)$
\State \Return $(\mathbf S_{\mathcal C},\mathbf P)$
\end{algorithmic}
\end{algorithm}

\textbf{Pairmixer block.} \Cref{alg:pairmixer-block} updates only the pair representation using outgoing and incoming triangle multiplication followed by a gated pair transition. The mask $\mathbf m$ restricts all pair operations to valid atoms.

\begin{algorithm}[H]
\caption{Pairmixer block}
\label{alg:pairmixer-block}
\begin{algorithmic}[1]
\Require Pair representation $\mathbf P$ and valid-atom mask $\mathbf m$
\Ensure Updated pair representation $\mathbf P$
\State $M_{ij}\gets m_i m_j$
\AlgPhase{1. Aggregate information over outgoing and incoming triangles}
\Statex $\displaystyle
\operatorname{Contract}_{\mathrm{out}}(\mathbf A,\mathbf B)_{ij}
\gets\sum_k\mathbf A_{ik}\odot\mathbf B_{jk},\qquad
\operatorname{Contract}_{\mathrm{in}}(\mathbf A,\mathbf B)_{ij}
\gets\sum_k\mathbf A_{ki}\odot\mathbf B_{kj}$
\For{$d$ in the ordered sequence $(\mathrm{out},\mathrm{in})$}
    \State $\overline{\mathbf P}\gets\operatorname{Norm}_{d}(\mathbf P)$
    \Statex $\displaystyle
    [\mathbf A,\mathbf B]\gets\operatorname{Split}\!\left(
    W_p^{d}\overline{\mathbf P}
    \odot\sigma(W_g^{d}\overline{\mathbf P})\right)\odot\mathbf M$
    \State $\mathbf U\gets\operatorname{Contract}_{d}(\mathbf A,\mathbf B)$
    \Statex $\displaystyle
    \mathbf P\gets\mathbf P+
    W_o^{d}\operatorname{Norm}'_{d}(\mathbf U)
    \odot\sigma(W_{go}^{d}\overline{\mathbf P})$
\EndFor
\Statex
\AlgPhase{2. Mix feature channels within each pair}
\State $\overline{\mathbf P}\gets\operatorname{Norm}_{\mathrm{tr}}(\mathbf P)$
\Statex $\displaystyle
\mathbf P\gets\mathbf P+W_{\mathrm{tr}}\!\left[
\operatorname{SiLU}(W_a\overline{\mathbf P})
\odot W_b\overline{\mathbf P}\right]$
\State \Return $\mathbf P$
\end{algorithmic}
\end{algorithm}

\textbf{Noisy-state and time embedding.} In \Cref{alg:noisy-state-embedding}, $\boldsymbol\Omega$ and $\boldsymbol\eta$ are fixed random Fourier frequencies and phases for encoding the coordinates, while $\boldsymbol\omega$ contains the fixed frequencies used for the sinusoidal time embedding.

\begin{algorithm}[H]
\caption{Noisy-state and time embedding}
\label{alg:noisy-state-embedding}
\begin{algorithmic}[1]
\Require Condition-only single representation $\mathbf S_{\mathcal C}$,
noisy coordinates $\mathbf X_t$, noisy lattice $\boldsymbol\ell_t$, and time $t$
\Ensure DiT input $\mathbf S$ and time condition $\mathbf c_t$
\AlgPhase{1. Embed noisy coordinates into the single track}
\State $\boldsymbol\Phi_i
    \gets\sqrt{2}\cos(\boldsymbol\Omega\odot\mathbf X_{t,i}
    +\boldsymbol\eta)$
    \Comment{fixed random Fourier features}
\State $\mathbf e_i^X\gets
    \operatorname{MLP}_X(\operatorname{vec}(\boldsymbol\Phi_i))$
\Statex
\AlgPhase{2. Embed and broadcast the noisy lattice}
\State $\mathbf e^L\gets\operatorname{MLP}_L(\boldsymbol\ell_t)$
\State $\mathbf S_i\gets
    \mathbf S_{\mathcal C,i}+\mathbf e_i^X+\mathbf e^L$
    \Comment{$\mathbf e^L$ is broadcast over atoms}
\Statex
\AlgPhase{3. Embed flow time}
\State $\boldsymbol\psi(t)\gets
    [\cos(\boldsymbol\omega t),\sin(\boldsymbol\omega t)]$
    \Comment{fixed sinusoidal frequencies}
\State $\mathbf c_t\gets\operatorname{MLP}_t(\boldsymbol\psi(t))$
\State \Return $(\mathbf S,\mathbf c_t)$
\end{algorithmic}
\end{algorithm}

\textbf{DiT block.} In \Cref{alg:dit-block}, $\boldsymbol\beta$, $\boldsymbol\alpha$, and $\boldsymbol\gamma$ denote the adaptive shift, scale, and residual-gate vectors, respectively. Subscripts $A$ and $F$ distinguish modulation parameters for the attention and feed-forward updates.

\begin{algorithm}[H]
\caption{Pair-biased DiT block}
\label{alg:dit-block}
\begin{algorithmic}[1]
\Require Single representation $\mathbf S$, time condition $\mathbf c_t$,
refined pair representation $\mathbf P^\star$, and mask $\mathbf m$
\Ensure Updated single representation $\mathbf S$
\AlgPhase{1. Generate time-dependent modulation}
\Statex $\displaystyle
[\boldsymbol\beta_A,\boldsymbol\alpha_A,\boldsymbol\gamma_A,
  \boldsymbol\beta_F,\boldsymbol\alpha_F,\boldsymbol\gamma_F]
\gets W_c\phi(\mathbf c_t)$
\State $\mathbf U_i\gets
    (1+\boldsymbol\alpha_A)\odot\operatorname{Norm}_1(\mathbf S_i)
    +\boldsymbol\beta_A$
\State $\widetilde{\mathbf U}_i\gets\operatorname{Norm}_A(\mathbf U_i)$
\Statex
\AlgPhase{2. Apply self-attention with pair bias}
\State $\mathbf q_i^h\gets
    \operatorname{Norm}_q(W_q^h\widetilde{\mathbf U}_i)$,
    $\mathbf k_i^h\gets
    \operatorname{Norm}_k(W_k^h\widetilde{\mathbf U}_i)$,
    $\mathbf v_i^h\gets W_v^h\widetilde{\mathbf U}_i$
\State $b_{ij}^h\gets W_P^h
    \operatorname{Norm}_P(\mathbf P_{ij}^\star)$
\Statex $\displaystyle
a_{ij}^h\gets\operatorname{softmax}_{j}\!\left(
\frac{\langle\mathbf q_i^h,\mathbf k_j^h\rangle}{\sqrt{d_h}}
+b_{ij}^h-\infty(1-m_j)\right)$
\State $\mathbf o_i^h\gets\sum_j a_{ij}^h\mathbf v_j^h$
\Statex $\displaystyle
\mathbf o_i\gets W_o\!\left[
\sigma(W_g\widetilde{\mathbf U}_i)
\odot\operatorname{Concat}_h(\mathbf o_i^h)\right]$
\State $\mathbf S_i\gets
    \mathbf S_i+\boldsymbol\gamma_A\odot\mathbf o_i$
\Statex
\AlgPhase{3. Apply the gated feed-forward update}
\State $\mathbf H_i\gets
    (1+\boldsymbol\alpha_F)\odot\operatorname{Norm}_2(\mathbf S_i)
    +\boldsymbol\beta_F$
\Statex $\displaystyle
\mathbf f_i\gets W_{\mathrm{down}}\!\left[
\phi(W_{\mathrm{gate}}\mathbf H_i)
\odot W_{\mathrm{up}}\mathbf H_i\right]$
\State $\mathbf S_i\gets
    \mathbf S_i+\boldsymbol\gamma_F\odot\mathbf f_i$
\State \Return $\mathbf S$
\end{algorithmic}
\end{algorithm}

\textbf{Prediction heads.} \Cref{alg:prediction-heads} maps the final single representation to centered Cartesian coordinates and a pooled lattice prediction over valid atoms.

\begin{algorithm}[H]
\caption{Coordinate and lattice prediction heads}
\label{alg:prediction-heads}
\begin{algorithmic}[1]
\Require Final single representation $\mathbf S$ and valid-atom mask $\mathbf m$
\Ensure Coordinate endpoint $\widehat{\mathbf X}_1$ and lattice endpoint
$\widehat{\boldsymbol\ell}_1$
\State $N\gets\sum_i m_i$
\AlgPhase{1. Predict centered atomic coordinates}
\State $\mathbf r_i\gets
    \operatorname{Linear}_X(\operatorname{Norm}_X(\mathbf S_i))$
\State $\displaystyle
    \widehat{\mathbf X}_{1,i}\gets\mathbf r_i-
    \frac{1}{N}\sum_jm_j\mathbf r_j$
\Statex
\AlgPhase{2. Pool atom features and predict the lattice}
\State $\displaystyle
    \overline{\mathbf s}\gets\frac{1}{N}\sum_i m_i
    \operatorname{Norm}_L(\mathbf S_i)$
\State $\widehat{\boldsymbol\ell}_1\gets
    \operatorname{MLP}_L(\overline{\mathbf s})$
\State \Return $(\widehat{\mathbf X}_1,\widehat{\boldsymbol\ell}_1)$
\end{algorithmic}
\end{algorithm}

\clearpage
\section{Dataset Statistics}
\label{app:dataset-statistics}

We report statistics for the CLARI training and validation splits used in our benchmark experiments (\Cref{sec:benchmarks}). Our preprocessing adds missing hydrogens using the CSD Python API~\citep{sykes2024scripting} before applying the 512-atom limit. This reduces the training split from 917{,}014 to 912{,}807 structures and the validation split from 1{,}048 to 1{,}047. We report both statistics for completeness.

\begin{table}[H]
\centering
\caption{\textbf{Size statistics for the CLARI training and validation splits.} Statistics are reported before and after applying the 512-atom unit-cell limit following hydrogen completion.}
\label{tab:clari-split-statistics}
\begin{tabular}{@{}lrrrrrrrr@{}}
\toprule
& \multicolumn{4}{c}{Unfiltered} & \multicolumn{4}{c}{$\leq512$ atoms} \\
\cmidrule(lr){2-5}\cmidrule(l){6-9}
Property & Min & Mean & Median & Max & Min & Mean & Median & Max \\
\midrule
\multicolumn{9}{@{}l}{\textbf{Training}} \\
\addlinespace[2pt]
Atoms / unit cell      & 5    & 203.65 & 180   & 3{,}558 & 5    & 201.71 & 180   & 512 \\
Molecules / unit cell  & 1    & 5.42   & 4     & 246     & 1    & 5.35   & 4     & 160 \\
Atoms / molecule       & 1.06 & 48.96  & 42.00 & 682.00  & 1.06 & 48.89  & 42.00 & 512.00 \\
Formula-unit count $Z$ & 1    & 3.34   & 4     & 40      & 1    & 3.33   & 4     & 40 \\
\addlinespace
\multicolumn{9}{@{}l}{\textbf{Validation}} \\
\addlinespace[2pt]
Atoms / unit cell      & 10   & 205.95 & 185   & 617    & 10   & 205.56 & 184   & 512 \\
Molecules / unit cell  & 1    & 5.51   & 4     & 99     & 1    & 5.42   & 4     & 52 \\
Atoms / molecule       & 5.00 & 48.42  & 42.10 & 242.00 & 5.00 & 48.46  & 42.20 & 242.00 \\
Formula-unit count $Z$ & 1    & 3.39   & 4     & 16     & 1    & 3.39   & 4     & 16 \\
\bottomrule
\end{tabular}
\end{table}

We additionally report chemical composition, molecular flexibility, and the availability of optional conditioning information. Similar to \citet{jin2025oxtal}, we label a structure as rigid if every molecular component contains at most three rotatable bonds and as flexible if any component contains more than three. This provides a simple heuristic rather than a complete description of molecular flexibility. \Cref{tab:clari-composition} summarizes these properties before and after the 512-atom filtering.

\begin{table}[H]
\centering
\caption{\textbf{Composition, flexibility, and conditioning availability in the CLARI splits.} Entries report the number of structures before and after applying the 512-atom unit-cell limit.}
\label{tab:clari-composition}
\begin{tabular}{@{}lcccc@{}}
\toprule
& \multicolumn{2}{c}{Training} & \multicolumn{2}{c}{Validation} \\
\cmidrule(lr){2-3}\cmidrule(l){4-5}
Property & Unfiltered & $\leq512$ atoms & Unfiltered & $\leq512$ atoms \\
\midrule
\multicolumn{5}{@{}l}{\textbf{\emph{Composition}}} \\
\quad Organic
& 491{,}219
& 490{,}475
& 573
& 573 \\
\quad Organometallic
& 425{,}795
& 422{,}332
& 475
& 474 \\
\addlinespace
\multicolumn{5}{@{}l}{\textbf{\emph{Molecular flexibility}}} \\
\quad Rigid
& 194{,}017
& 193{,}620
& 226
& 226 \\
\quad Flexible
& 722{,}997
& 719{,}187
& 822
& 821 \\
\addlinespace
\multicolumn{5}{@{}l}{\textbf{\emph{Conditioning availability}}} \\
\quad Template
& 598{,}141
& 596{,}447
& 683
& 683 \\
\quad Stereochemistry
& 917{,}014
& 912{,}807
& 1{,}048
& 1{,}047 \\
\quad Space group
& 917{,}012
& 912{,}805
& 1{,}048
& 1{,}047 \\
\bottomrule
\end{tabular}
\end{table}

\clearpage
\section{Training and Sampling Hyperparameters}
\label{app:training-sampling-hyperparameters}

We summarize the hyperparameters used for \model{}-M and \model{}-L.
Unless noted, all settings are shared between the two model sizes.

\textbf{Architecture.}
\Cref{tab:ours-architecture-hyperparameters} specifies model capacity and
backbone configuration; the two models differ only in width, head count, and
parameter count.

\begin{table}[H]
\centering
\caption{\textbf{Architecture hyperparameters.}
Model capacity and backbone configuration for \ours{}-M and \ours{}-L.}
\label{tab:ours-architecture-hyperparameters}
\begin{tabular}{@{}lrr@{}}
\toprule
Parameter & \ours{}-M & \ours{}-L \\
\midrule
\multicolumn{3}{@{}l}{\textbf{\emph{Capacity}}} \\
\quad Parameters       & 88M & 187M \\
\quad Single width     & 512 & 768 \\
\quad Pair width       & 128 & 192 \\
\quad Attention heads  & 8   & 12 \\
\addlinespace
\multicolumn{3}{@{}l}{\textbf{\emph{Backbone}}} \\
\quad Pairmixer blocks & 4 & 4 \\
\quad DiT blocks       & 16 & 16 \\
\quad MLP ratio        & 4 & 4 \\
\quad QK normalization & On & On \\
\quad Feed-forward network & SwiGLU & SwiGLU \\
\quad Dropout          & 0 & 0 \\
\bottomrule
\end{tabular}
\end{table}

\textbf{Training.}
\Cref{tab:ours-training-hyperparameters} summarizes the flow-matching,
optimization, conditioning, and checkpoint-selection settings.

\begin{table}[H]
\centering
\caption{\textbf{Training hyperparameters.}
Flow-matching, optimization, conditioning, and model-selection settings.
All settings are shared between \ours{}-M and \ours{}-L unless noted.}
\label{tab:ours-training-hyperparameters}
\begin{tabular}{@{}>{\raggedright\arraybackslash}p{0.39\columnwidth}
                    >{\raggedright\arraybackslash}p{0.55\columnwidth}@{}}
\toprule
Parameter & Value \\
\midrule
\multicolumn{2}{@{}l}{\textbf{\emph{Flow matching}}} \\
\quad Priors & Centered Gaussian fpr $\mathbf{X}$; standard Gaussian for $\boldsymbol{\ell}$ \\
\quad Interpolant & Linear ($x_t=(1-t)x_0+tx_1$) \\
\quad Time distribution &
$0.98\,\operatorname{Beta}(1.9,1.0)+0.02\,\operatorname{Uniform}(0,1)$ \\
\quad Objective & $L^1$ coordinate--lattice loss \\
\quad Loss weights & Coordinates 10; lattice 1 \\
\addlinespace
\multicolumn{2}{@{}l}{\textbf{\emph{Optimization}}} \\
\quad Optimizer & Muon for hidden matrices; AdamW otherwise \\
\quad Learning rate & $10^{-4}$ \\
\quad Weight decay & 0 (\ours{}-M); $10^{-2}$ (\ours{}-L) \\
\quad Muon settings & Momentum 0.95; Nesterov; 5 Newton--Schulz steps \\
\quad AdamW settings & $\beta_1=0.9$, $\beta_2=0.999$, $\epsilon=10^{-8}$ \\
\quad Schedule & 10{,}000-step linear warmup from zero \\
\quad Gradient clipping & Global norm, 1.0 \\
\quad EMA decay & 0.9999 \\
\addlinespace
\multicolumn{2}{@{}l}{\textbf{\emph{Conditioning and augmentation}}} \\
\quad Condition dropout & Template 0.5; stereochemistry 0.5; space group 0.9 \\
\quad Augmentation & Crystal translation; template rotation and translation \\
\addlinespace
\multicolumn{2}{@{}l}{\textbf{\emph{Runtime and selection}}} \\
\quad Precision & bfloat16 mixed precision \\
\quad Checkpoint interval & Every 50 epochs \\
\quad Checkpoint selection & Highest validation $\mathrm{Sol}_C$ \\
\bottomrule
\end{tabular}
\end{table}

\textbf{Curriculum and batching.}
We use two-stage training with fixed-shape batches to enable
\texttt{torch.compile}. Each structure is assigned to the smallest
atom-count bucket that contains it and padded to that bucket size.

\begin{table}[H]
\centering
\caption{\textbf{Two-stage training curriculum and batching.}
Atom-count buckets and batch sizes used for fixed-shape training across eight GPUs.}
\label{tab:ours-curriculum-batches}
\begin{tabular}{@{}>{\raggedright\arraybackslash}p{0.39\columnwidth}
                    >{\raggedright\arraybackslash}p{0.55\columnwidth}@{}}
\toprule
Parameter & Value \\
\midrule
\multicolumn{2}{@{}l}{\textbf{\emph{Stage 1}}} \\
\quad Maximum atoms & 300 \\
\quad Bucket caps & 32, 64, 96, 128, 160, 192, 224, 256, 288, 300 \\
\quad Per-GPU batch & 16 for every bucket \\
\quad Global batch & 128 \\
\addlinespace
\multicolumn{2}{@{}l}{\textbf{\emph{Stage 2}}} \\
\quad Maximum atoms & 512 \\
\quad Additional bucket caps & 320, 352, 384, 416, 448, 480, 512 \\
\quad Per-GPU batch & 16 ($\leq300$), 14 (320), 11 (352), 9 (384),
8 (416), 7 (448), 6 (480), 5 (512) \\
\quad Global batch & 128, 112, 88, 72, 64, 56, 48, and 40, respectively \\
\bottomrule
\end{tabular}
\end{table}

\textbf{Sampling.}
Both model sizes use the same sampling and conditioning configuration for
benchmark generation.

\begin{table}[H]
\centering
\caption{\textbf{Sampling hyperparameters.}
Sampling and conditioning configuration used for benchmark generation.}
\label{tab:ours-sampling-hyperparameters}
\begin{tabular}{@{}>{\raggedright\arraybackslash}p{0.39\columnwidth}
                    >{\raggedright\arraybackslash}p{0.55\columnwidth}@{}}
\toprule
Parameter & Value \\
\midrule
\multicolumn{2}{@{}l}{\textbf{\emph{Numerical solver}}} \\
\quad Solver & EDM--Heun with stochastic churn \\
\quad Noise schedule & Karras \\
\quad Sampling steps & 200 \\
\quad Sigma range & $\sigma_{\min}=0.002$, $\sigma_{\max}=80$ \\
\quad Schedule exponent & $\rho=7$ \\
\quad Churn settings & $S_{\mathrm{churn}}=60$, $S_{\min}=0$,
$S_{\max}=999$, $S_{\mathrm{noise}}=1.003$ \\
\addlinespace
\multicolumn{2}{@{}l}{\textbf{\emph{Generation protocol}}} \\
\quad Weights & EMA \\
\quad Conditioning & Template and available stereochemistry on; space group off \\
\quad Template augmentation & Random rotation and translation \\
\quad Candidates per target & 1{,}000 \\
\quad Random seed & 42 \\
\bottomrule
\end{tabular}
\end{table}

\clearpage
\section{Structure Generation Evaluation}
\label{app:structure-generation-evaluation}

This section details the evaluation protocol for structure generation. We
formalize the crystal solve rate $\mathrm{Sol}_C$~\citep{jin2025oxtal},
including COMPACK~\citep{chisholm2005compack} matching, collision detection,
and experimental-reference selection. We also summarize the differences between
the released OXtal and CLARI evaluators that motivate our re-evaluation of CLARI.

\textbf{Definition of crystal solve rate.}
Let $c\in\{1,\ldots,C\}$ index targets and
$s\in\{1,\ldots,n_c\}$ index generated candidates for target $c$. The
crystal-level solve rate is
\begin{equation}
    \mathrm{Sol}_C
    =\frac{1}{C}\sum_{c=1}^{C}\max_{1\leq s\leq n_c}
        \mathrm{sol}_{c,s},
\end{equation}
where $\mathrm{sol}_{c,s}$ indicates whether candidate $x_{c,s}$ solves target
$c$:
\begin{equation}
    \mathrm{sol}_{c,s}
    =\mathbf{1}\!\left[
        \mathrm{col}_{c,s}=0
        \ \land\ n_{15}(x_{c,s},t^\star_{c,s})\geq 8
        \ \land\ r_{15}(x_{c,s},t^\star_{c,s})<2\,\text{\AA}
    \right],
\end{equation}
where $\mathrm{col}_{c,s}$ is the collision indicator; $n_{15}$ and $r_{15}$
denote the number of matched molecules and RMSD, respectively, returned by
COMPACK at packing size 15; and $t^\star_{c,s}$ is the selected experimental
reference.

Intuitively, a candidate is collision-free ($\mathrm{col}_{c,s}=0$) if all
nonbonded heavy-atom pairs are separated beyond the collision threshold.
Specifically,
\begin{equation}
    \mathrm{col}_{c,s}
    =\mathbf{1}\!\left[
        \exists(i,j):
        r_i^{\mathrm{vdW}}+r_j^{\mathrm{vdW}}-d_{ij}^{(c,s)}
        \geq 0.7\,\text{\AA}
    \right],
\end{equation}
where $d_{ij}^{(c,s)}$ is the distance between a nonbonded heavy-atom pair
$(i,j)$, and $r_i^{\mathrm{vdW}}$ and $r_j^{\mathrm{vdW}}$ are their van der
Waals radii.

For targets with multiple experimental references, we select the best-matching
reference as $t^\star_{c,s}$. Let $T(c)$ denote the reference set for target $c$
and define
\begin{equation}
    P_{c,s}=\{t\in T(c):n_{15}(x_{c,s},t)\geq 8\}.
\end{equation}
OXtal selects
\begin{equation}
    t^\star_{c,s}=
    \begin{cases}
        \displaystyle\arg\min_{t\in P_{c,s}} r_{15}(x_{c,s},t),
        & P_{c,s}\neq\varnothing,\\[3pt]
        \displaystyle\arg\max_{t\in T(c)} n_{15}(x_{c,s},t),
        & P_{c,s}=\varnothing.
    \end{cases}
\end{equation}

\textbf{Differences between the OXtal and CLARI evaluators.}
The released CLARI evaluator differs from the OXtal evaluator in two aspects:
(1) collision detection, including the treatment of intermolecular collisions,
and (2) selection of the best experimental reference for targets with multiple
reference structures. For a consistent comparison, we therefore re-evaluate
CLARI-M, CLARI-L, and CLARI-H using the OXtal definitions.
\clearpage
\section{Detailed Structure-Generation Results}
\label{app:structure-generation-results}

This appendix provides tabular results for the structure-generation benchmarks (\Cref{sec:benchmarks}) across candidate budgets and evaluation criteria.

\Cref{tab:benchmark_solc_main} reports the standard $\mathrm{Sol}_C$ and strict
$\mathrm{Sol}_C^{15/15}$ at budgets of 30 and 1{,}000 candidates, with
resampling standard deviations reported at the 30-candidate budget.
\Cref{tab:combined_all_metrics} provides additional collision, packing, and
recovery metrics for \model{}-M and \model{}-L at the matched 30-candidate
budget.

\begin{table}[H]
\centering
\caption{\textbf{Tabular results for structure generation benchmark.}
We report standard $\mathrm{Sol}_C$ and strict $\mathrm{Sol}_C^{15/15}$ at
30 and 1{,}000 candidates per target. At 30 candidates, CLARI and \model{}
results are mean $\pm$ standard deviation over 5{,}000 resamples from
1{,}000-candidate pools; other baselines are reported as provided in \citet{jin2025oxtal}.
Results at 1{,}000 candidates use the full pool.
\textbf{Bold} and \underline{underline} denote the best and second-best
generative results, respectively.}
\label{tab:benchmark_solc_main}
\small
\setlength{\tabcolsep}{3.5pt}
\begin{tabular}{lcccccc}
\toprule
Method & Rigid (50) & Flexible (50) & CSP5 (6) & CSP6 (5) & CSP7 (8) & Teaching (773) \\

\midrule
\multicolumn{7}{l}{\textbf{\emph{$\mathrm{Sol}_C$ at 30-candidate budget}}} \\
\addlinespace
A-Transformer & 0.060 & 0.000 & 0.000 & 0.000 & 0.000 & -- \\
AssembleFlow & 0.000 & 0.000 & 0.000 & 0.000 & 0.000 & -- \\
OXtal & 0.300 & 0.220 & 0.167 & 0.200 & 0.125 & -- \\
CLARI-M & $0.692 \pm 0.038$ & $0.283 \pm 0.044$ & $0.410 \pm 0.136$ & $0.286 \pm 0.119$ & $0.154 \pm 0.102$ & $0.506 \pm 0.010$ \\
CLARI-L & $0.728 \pm 0.039$ & $0.336 \pm 0.043$ & $0.541 \pm 0.160$ & $0.332 \pm 0.138$ & $0.138 \pm 0.105$ & $0.549 \pm 0.010$ \\
CLARI-H & $0.740 \pm 0.039$ & $0.396 \pm 0.042$ & $0.598 \pm 0.146$ & $0.412 \pm 0.158$ & $0.134 \pm 0.112$ & $0.590 \pm 0.010$ \\
\rowcolor{mfTealFill}
\model{}-M & $\mathbf{0.807} \pm 0.031$ & $\underline{0.498} \pm 0.047$ & $\underline{0.732} \pm 0.130$ & $\underline{0.426} \pm 0.154$ & $\underline{0.212} \pm 0.102$ & $\underline{0.637} \pm 0.009$ \\
\rowcolor{mfTealFill}
\model{}-L & $\underline{0.795} \pm 0.035$ & $\mathbf{0.530} \pm 0.047$ & $\mathbf{0.793} \pm 0.081$ & $\mathbf{0.433} \pm 0.165$ & $\mathbf{0.238} \pm 0.126$ & $\mathbf{0.663} \pm 0.009$ \\
\midrule
\multicolumn{7}{l}{\textbf{\emph{$\mathrm{Sol}_C^{15/15}$ at 30-candidate budget}}} \\
\addlinespace
CLARI-M & $0.264 \pm 0.038$ & $0.041 \pm 0.025$ & $0.069 \pm 0.091$ & $0.006 \pm 0.033$ & $0.010 \pm 0.034$ & $0.165 \pm 0.008$ \\
CLARI-L & $0.351 \pm 0.041$ & $0.107 \pm 0.037$ & $0.131 \pm 0.100$ & $0.063 \pm 0.102$ & $\underline{0.018} \pm 0.043$ & $0.223 \pm 0.009$ \\
CLARI-H & $0.424 \pm 0.043$ & $0.155 \pm 0.039$ & $0.230 \pm 0.124$ & $0.088 \pm 0.122$ & $0.011 \pm 0.036$ & $0.275 \pm 0.009$ \\
\rowcolor{mfTealFill}
\model{}-M & $\mathbf{0.508} \pm 0.039$ & $\underline{0.262} \pm 0.043$ & $\underline{0.374} \pm 0.149$ & $\underline{0.108} \pm 0.127$ & $0.011 \pm 0.036$ & $\underline{0.367} \pm 0.009$ \\
\rowcolor{mfTealFill}
\model{}-L & $\underline{0.476} \pm 0.045$ & $\mathbf{0.296} \pm 0.043$ & $\mathbf{0.489} \pm 0.172$ & $\mathbf{0.254} \pm 0.128$ & $\mathbf{0.025} \pm 0.053$ & $\mathbf{0.394} \pm 0.009$ \\
\midrule
\multicolumn{7}{l}{\textbf{\emph{$\mathrm{Sol}_C$ at 1{,}000-candidate budget}}} \\
\addlinespace
CLARI-M & \underline{0.960} & 0.740 & \underline{0.833} & \underline{0.800} & 0.625 & 0.811 \\
CLARI-L & \underline{0.960} & 0.760 & \textbf{1.000} & \underline{0.800} & \underline{0.875} & 0.841 \\
CLARI-H & 0.940 & 0.800 & \textbf{1.000} & \underline{0.800} & \underline{0.875} & 0.869 \\
\rowcolor{mfTealFill}
\model{}-M & \underline{0.960} & \textbf{0.920} & \textbf{1.000} & \textbf{1.000} & \textbf{1.000} & \underline{0.895} \\
\rowcolor{mfTealFill}
\model{}-L & \textbf{0.980} & \underline{0.880} & \textbf{1.000} & \textbf{1.000} & \textbf{1.000} & \textbf{0.900} \\
\midrule
\multicolumn{7}{l}{\textbf{\emph{$\mathrm{Sol}_C^{15/15}$ at 1{,}000-candidate budget}}} \\
\addlinespace
CLARI-M & 0.680 & 0.380 & 0.333 & 0.200 & 0.125 & 0.453 \\
CLARI-L & 0.780 & 0.540 & 0.500 & \underline{0.800} & 0.125 & 0.563 \\
CLARI-H & \underline{0.800} & 0.640 & \underline{0.667} & \underline{0.800} & \underline{0.250} & 0.618 \\
\rowcolor{mfTealFill}
\model{}-M & \underline{0.800} & \textbf{0.740} & \textbf{0.833} & \underline{0.800} & \underline{0.250} & \underline{0.683} \\
\rowcolor{mfTealFill}
\model{}-L & \textbf{0.900} & \underline{0.700} & \textbf{0.833} & \textbf{1.000} & \textbf{0.500} & \textbf{0.722} \\
\bottomrule
\end{tabular}
\vspace{1mm}
\begin{minipage}{0.98\textwidth}
\end{minipage}
\end{table}

\begin{table}[H]
\centering
\caption{\textbf{Full metrics at a 30-candidate budget.}
We report per-sample collision, packing, and recovery rates
($\mathrm{Col}_S$, $\mathrm{Pac}_S$, $\mathrm{Rec}_S$) and crystal-level
packing, recovery, and solve rates
($\mathrm{Pac}_C$, $\mathrm{Rec}_C$, $\mathrm{Sol}_C$), following the
definitions in \citet{jin2025oxtal}. Other baselines use their exact
30-candidate pools, while \model{}-M and \model{}-L report means over 5{,}000
seeded draws of 30 candidates from their 1{,}000-candidate pools. CLARI is
omitted because only $\mathrm{Sol}_C$ is available.}
\label{tab:combined_all_metrics}
\begin{tabular}{llcccccc}
\toprule
Dataset & Method & $\mathrm{Col}_S \downarrow$ & $\mathrm{Pac}_S \uparrow$ &
$\mathrm{Pac}_C \uparrow$ &
$\mathrm{Rec}_S \uparrow$ & $\mathrm{Rec}_C \uparrow$ &
$\mathrm{Sol}_C \uparrow$ \\
\midrule
Rigid & A-Transformer & 0.731 & 0.015 & 0.060 & 0.033 & 0.120 & 0.060 \\
 & AssembleFlow & 0.524 & 0.001 & 0.040 & 0.211 & 0.760 & 0.000 \\
 & OXtal & 0.011 & 0.873 & 1.000 & 0.737 & 0.960 & 0.300 \\
\rowcolor{mfTealFill}
 & \model{}-M & 0.036 & 0.433 & 0.876 & 0.860 & 0.978 & 0.807 \\
\rowcolor{mfTealFill}
 & \model{}-L & 0.028 & 0.405 & 0.872 & 0.847 & 0.979 & 0.795 \\
\midrule
Flexible & A-Transformer & 0.900 & 0.001 & 0.020 & 0.000 & 0.000 & 0.000 \\
 & AssembleFlow & 0.883 & 0.000 & 0.000 & 0.021 & 0.140 & 0.000 \\
 & OXtal & 0.097 & 0.291 & 0.900 & 0.048 & 0.400 & 0.220 \\
\rowcolor{mfTealFill}
 & \model{}-M & 0.030 & 0.099 & 0.578 & 0.346 & 0.879 & 0.498 \\
\rowcolor{mfTealFill}
 & \model{}-L & 0.031 & 0.099 & 0.605 & 0.371 & 0.920 & 0.530 \\
\midrule
CSP5 & A-Transformer & 0.833 & 0.000 & 0.000 & 0.000 & 0.000 & 0.000 \\
 & AssembleFlow & 0.717 & 0.000 & 0.000 & 0.150 & 0.500 & 0.000 \\
 & OXtal & 0.006 & 0.667 & 0.833 & 0.572 & 0.833 & 0.167 \\
\rowcolor{mfTealFill}
 & \model{}-M & 0.023 & 0.251 & 0.808 & 0.784 & 0.915 & 0.732 \\
\rowcolor{mfTealFill}
 & \model{}-L & 0.018 & 0.328 & 0.797 & 0.792 & 0.902 & 0.793 \\
\midrule
CSP6 & A-Transformer & 0.967 & 0.000 & 0.000 & 0.000 & 0.000 & 0.000 \\
 & AssembleFlow & 0.800 & 0.000 & 0.000 & 0.073 & 0.200 & 0.000 \\
 & OXtal & 0.013 & 0.660 & 1.000 & 0.160 & 0.600 & 0.200 \\
\rowcolor{mfTealFill}
 & \model{}-M & 0.070 & 0.177 & 0.607 & 0.559 & 0.851 & 0.426 \\
\rowcolor{mfTealFill}
 & \model{}-L & 0.052 & 0.193 & 0.661 & 0.578 & 0.902 & 0.433 \\
\midrule
CSP7 & A-Transformer & 0.950 & 0.000 & 0.000 & 0.000 & 0.000 & 0.000 \\
 & AssembleFlow & 0.808 & 0.000 & 0.000 & 0.063 & 0.250 & 0.000 \\
 & OXtal & 0.021 & 0.483 & 0.875 & 0.129 & 0.375 & 0.125 \\
\rowcolor{mfTealFill}
 & \model{}-M & 0.178 & 0.041 & 0.396 & 0.251 & 0.656 & 0.212 \\
\rowcolor{mfTealFill}
 & \model{}-L & 0.174 & 0.035 & 0.469 & 0.278 & 0.681 & 0.238 \\
\midrule
\rowcolor{mfTealFill}
Teaching & \model{}-M & 0.057 & 0.281 & 0.704 & 0.536 & 0.842 & 0.637 \\
\rowcolor{mfTealFill}
 & \model{}-L & 0.050 & 0.288 & 0.730 & 0.554 & 0.859 & 0.663 \\
\bottomrule
\end{tabular}
\end{table}

\clearpage
\section{End-to-End Generation with Empirical \texorpdfstring{$Z$}{Z} Sampling}
\label{app:empirical-unit-cell-z}

The main structure-generation experiments in \Cref{sec:benchmarks} assume known formula-unit count $Z$ for comparison with CLARI. Here, we evaluate an end-to-end setting in which $Z$ is unknown and sampled from an empirical prior estimated from the training set.

\textbf{Prior construction.}
For each training crystal, we group identical molecular components by exact
component SMILES. Let $n_k$ denote the number of copies of distinct component
$G_k$ in the unit cell. We recover the formula-unit count $Z$ and reduced
stoichiometry $\mathbf r$ as
\begin{equation*}
Z=\gcd(n_1,\ldots,n_K),
\qquad
r_k=\frac{n_k}{Z}.
\end{equation*}
We apply this procedure to all 912{,}807 structures in the CLARI training split
after the 512-atom filter. If $c_z$ denotes the number of structures with
$Z=z$, the empirical prior is
\begin{equation*}
\widehat p_{\mathrm{train}}(Z=z)
=\frac{c_z}{\sum_{z'}c_{z'}}.
\end{equation*}
The resulting support is
$\{1,2,3,4,5,6,7,8,9,10,12,14,15,16,18,20,24,26,27,28,30,32,34,40\}$.

\textbf{Sampling formula-unit count $Z$.}
Let $N_{\mathrm{fu}}=\sum_k r_kN_k$ denote the number of atoms in one formula
unit for a benchmark target. To bound inference cost, we restrict $Z$ to values
that yield at most $N_{\max}=4096$ atoms and renormalize the empirical prior:
\begin{equation*}
\widehat p_{\mathrm{trunc}}(Z=z)
=
\frac{\widehat p_{\mathrm{train}}(Z=z)
\,\mathbf{1}[N_{\mathrm{fu}}z\leq N_{\max}]}
{\sum_{z'}\widehat p_{\mathrm{train}}(Z=z')
\,\mathbf{1}[N_{\mathrm{fu}}z'\leq N_{\max}]}.
\end{equation*}
For each candidate, we sample $Z$ from $\widehat p_{\mathrm{trunc}}$, replicate the formula unit accordingly, and pass the resulting unit-cell composition to \model{}. All other settings match the main benchmark.

\begin{figure}[H]
  \centering
  \includegraphics[width=\textwidth]{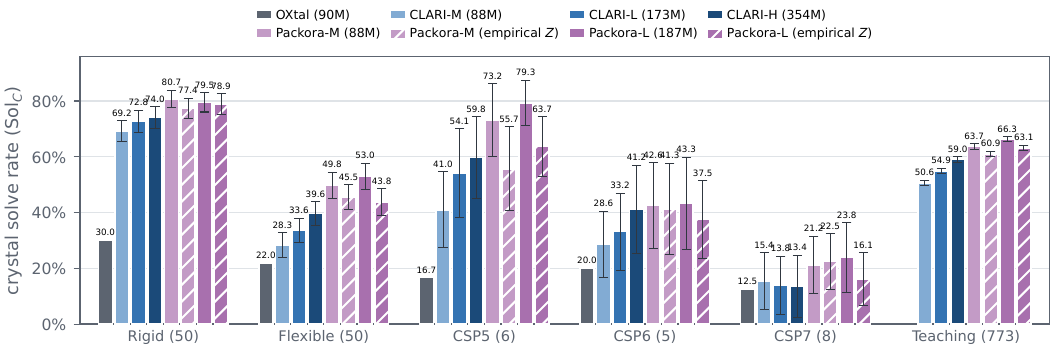}
  \caption{\textbf{Crystal coverage with empirical unit-cell $Z$ sampling at 30 candidates per target.}
  The solid bars reproduce \Cref{fig:benchmark-teaser}; diagonally hatched bars
  represent results with $Z$ sampled from an empirical prior. Bars report mean
  $\mathrm{Sol}_C$, and error bars show one standard deviation across 5{,}000
  resamples with replacement from each 1{,}000-candidate pool.}
  \label{fig:empirical-z-benchmark}
\end{figure}

\begin{figure}[H]
  \centering
  \includegraphics[width=\textwidth]{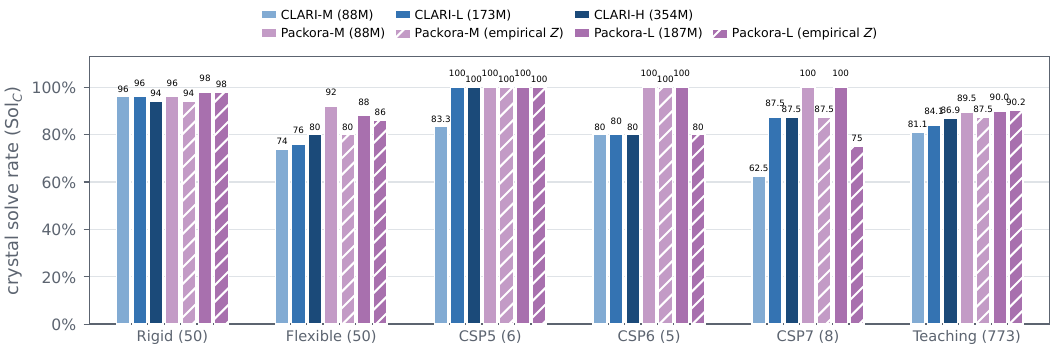}
  \caption{\textbf{Crystal coverage with empirical unit-cell $Z$ sampling at
  1{,}000 candidates per target.}
  The solid bars reproduce \Cref{fig:benchmark-exact}; diagonally hatched bars
  represent results with $Z$ sampled from an empirical prior. Bars report exact
  $\mathrm{Sol}_C$ over the full 1{,}000-candidate pools.}
  \label{fig:empirical-z-benchmark-exact}
\end{figure}

\begin{figure}[H]
  \centering
  \includegraphics[width=\textwidth]{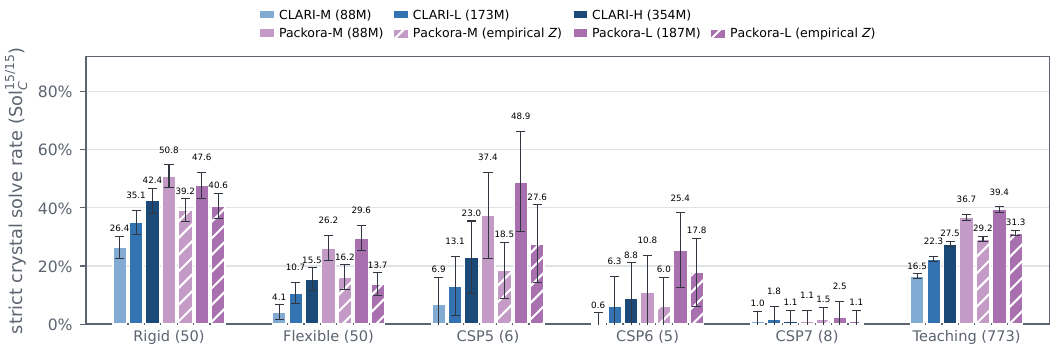}
  \caption{\textbf{Strict crystal coverage with empirical unit-cell $Z$ sampling
  at 30 candidates per target.}
  The solid bars reproduce \Cref{fig:benchmark-teaser-strict}; diagonally hatched
  bars represent results with $Z$ sampled from an empirical prior. Bars report
  mean $\mathrm{Sol}_C^{15/15}$, and error bars show one standard deviation
  across 5{,}000 resamples with replacement from each 1{,}000-candidate pool.}
  \label{fig:empirical-z-benchmark-strict}
\end{figure}

\begin{figure}[H]
  \centering
  \includegraphics[width=\textwidth]{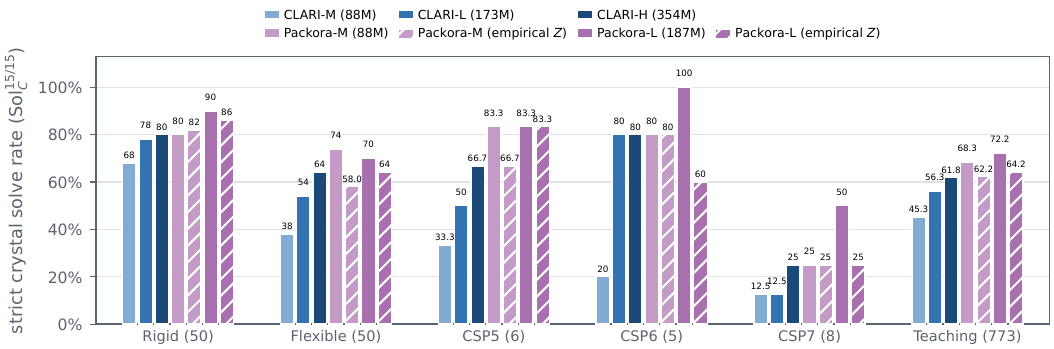}
  \caption{\textbf{Strict crystal coverage with empirical unit-cell $Z$ sampling
  at 1{,}000 candidates per target.}
  The solid bars reproduce \Cref{fig:benchmark-exact-strict}; diagonally hatched
  bars represent results with $Z$ sampled from an empirical prior. Bars report
  exact $\mathrm{Sol}_C^{15/15}$ over the full 1{,}000-candidate pools.}
  \label{fig:empirical-z-benchmark-exact-strict}
\end{figure}

\begin{figure}[H]
  \centering
  \includegraphics[width=\textwidth]{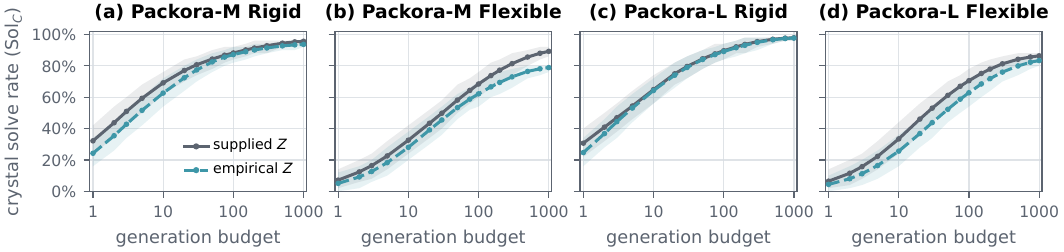}
  \caption{\textbf{Effect of empirical unit-cell $Z$ sampling across candidate budgets.}
  Curves show mean $\mathrm{Sol}_C$ for \model{}-M and \model{}-L on the Rigid
  and Flexible benchmarks over 5{,}000 resamples, and shaded bands show the
  2.5th--97.5th percentiles. The gray solid lines denote supplied $Z$, and the
  teal dotted lines denote empirical $Z$. The two settings remain close across
  candidate budgets.}
  \label{fig:empirical-z-candidate-budget}
\end{figure}

\textbf{Results.}
We evaluate \model{}-M and \model{}-L with empirical $Z$ sampling and find that
both remain competitive in this end-to-end setting.
\Cref{fig:empirical-z-benchmark,fig:empirical-z-benchmark-exact,fig:empirical-z-benchmark-strict,fig:empirical-z-benchmark-exact-strict}
compare empirical-$Z$ sampling with the main supplied-$Z$ setting at effective
30-candidate and exact 1{,}000-candidate budgets under both standard and strict
criteria. \Cref{fig:empirical-z-candidate-budget} shows how the
supplied and empirical performance gap changes with candidate budget on the Rigid and Flexible benchmarks.

\clearpage
\section{Structure Overlay Visualizations}
\label{app:strict-structure-overlays}

We visualize 18 representative strict matches by overlaying \model{}-L predictions with their experimental references. Experimental structures are shown as wider gray sticks and \model{}-L predictions as narrower green sticks; hydrogen atoms are omitted for clarity. The overlays are rendered with py3Dmol using the 3Dmol.js WebGL backend~\citep{rego2015_3dmol}.

\begin{figure}[H]
    \centering
    \includegraphics[width=0.93\textwidth,height=0.75\textheight,keepaspectratio]{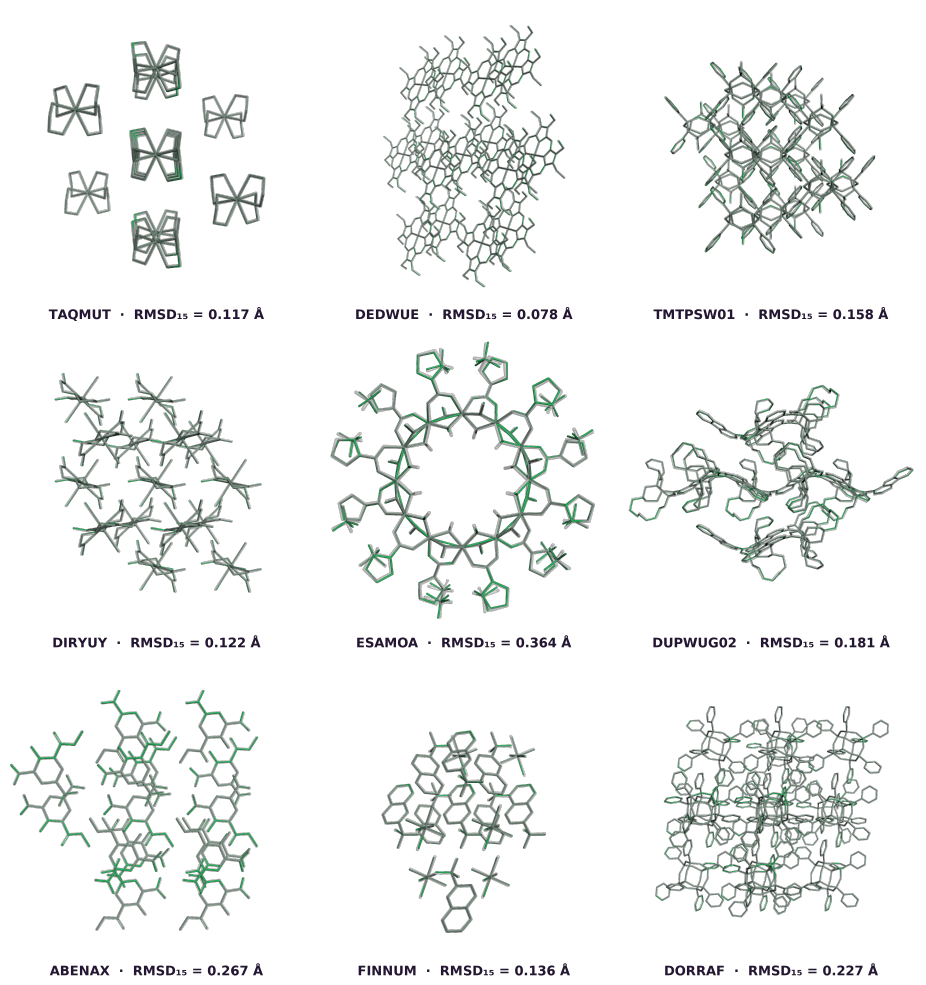}
    \caption{\textbf{Structure overlays for \model{}-L predictions.}
    Experimental structures are shown in gray and \model{}-L predictions in green.
    Labels report the CSD refcode and corresponding $\mathrm{RMSD}_{15}$.}
    \label{fig:strict-structure-overlays}
\end{figure}

\clearpage
\begin{figure}[H]
    \centering
    \includegraphics[width=0.93\textwidth,height=0.75\textheight,keepaspectratio]{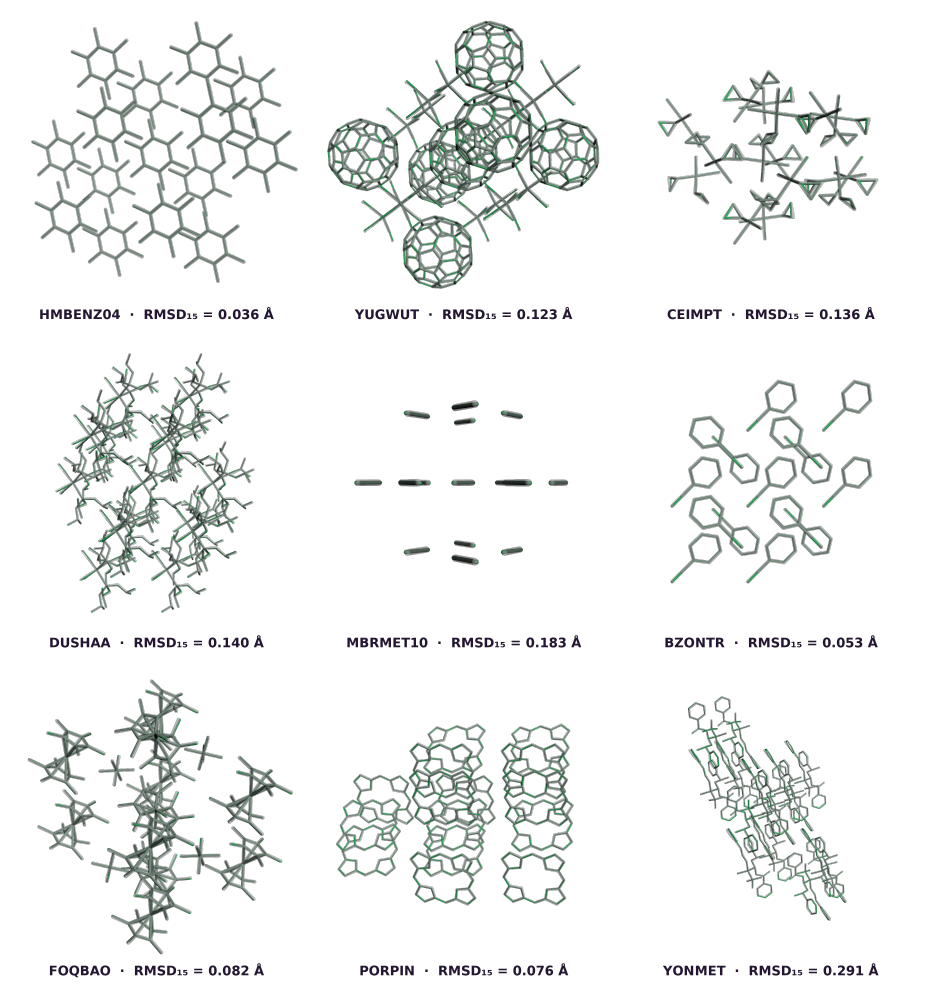}
    \caption{\textbf{Structure overlays for \model{}-L predictions.}
    Experimental structures are shown in gray and \model{}-L predictions in green.
    Labels report the CSD refcode and corresponding $\mathrm{RMSD}_{15}$.}
    \label{fig:strict-structure-overlays-additional}
\end{figure}
\clearpage
\section{Detailed Structure-Ranking Results}
\label{app:structure-ranking-results}

This appendix reports the complete tabular results underlying the FastCSP structure-ranking benchmarks in \Cref{sec:structure-ranking}. Each table gives the energy rank and $\mathrm{RMSD}_{30}$ of the best candidate matching each experimental form; NR denotes that no match was recovered.

\begin{table}[H]
\centering
\small
\caption{\textbf{Complete FastCSP single-polymorph results.}
Each method reports the energy rank and $\mathrm{RMSD}_{30}$ (\AA) of the
best candidate matching the experimental form. NR denotes no recovery, with
the corresponding $\mathrm{RMSD}_{30}$ unavailable.}
\label{tab:fastcsp-single-ranking-details}

\footnotesize
\setlength{\tabcolsep}{4pt}
\renewcommand{\arraystretch}{1.05}

\begin{tabular}{@{}llrrrrrr@{}}
\toprule
& & \multicolumn{2}{c}{CLARI-H}
& \multicolumn{2}{c}{\model{}-M}
& \multicolumn{2}{c}{\model{}-L} \\
\cmidrule(lr){3-4}\cmidrule(lr){5-6}\cmidrule(l){7-8}
Target & Subset
& Rank & $\mathrm{RMSD}_{30}$
& Rank & $\mathrm{RMSD}_{30}$
& Rank & $\mathrm{RMSD}_{30}$ \\
\midrule
Target II & Semi-rigid & 5 & 0.55 & NR & -- & 6 & 0.58 \\
Target IV & Semi-rigid & 1 & 0.16 & 1 & 0.14 & 1 & 0.18 \\
Target V & Semi-rigid & 8 & 0.33 & 5 & 0.18 & 7 & 0.12 \\
Target VIII & Semi-rigid & NR & -- & NR & -- & 1 & 0.22 \\
Target XII & Semi-rigid & 1 & 0.25 & 1 & 0.25 & 1 & 0.28 \\
Target XIII & Semi-rigid & 1 & 0.09 & 1 & 0.10 & 1 & 0.09 \\
Target XVI & Semi-rigid & NR & -- & NR & -- & NR & -- \\
Target XVII & Semi-rigid & 1 & 0.12 & 1 & 0.14 & 1 & 0.18 \\
Target XXII & Semi-rigid & 1 & 0.17 & 2 & 0.19 & 2 & 0.25 \\
Acetic Acid & Semi-rigid & NR & -- & NR & -- & NR & -- \\
Caprylolactam & Semi-rigid & NR & -- & 2 & 0.15 & 3 & 0.11 \\
CEBYUD & Semi-rigid & 1 & 0.14 & 1 & 0.14 & 1 & 0.14 \\
CUMJOJ & Semi-rigid & NR & -- & NR & -- & NR & -- \\
DEZDUH & Semi-rigid & 5 & 0.44 & 1 & 0.06 & 1 & 0.04 \\
Eniluracil & Semi-rigid & 5 & 0.63 & 1 & 0.11 & 1 & 0.10 \\
GACGAU & Semi-rigid & 2 & 0.12 & 2 & 0.12 & 2 & 0.17 \\
GOLHIB & Semi-rigid & NR & -- & 1 & 0.10 & 1 & 0.09 \\
HURYUQ & Semi-rigid & 27 & 0.13 & 5 & 0.09 & 1 & 0.15 \\
IHEPUG & Semi-rigid & NR & -- & NR & -- & NR & -- \\
LECZOL & Semi-rigid & NR & -- & NR & -- & NR & -- \\
ROHBUL & Semi-rigid & 1 & 0.08 & 1 & 0.05 & 1 & 0.04 \\
6-Fluorochromone & Semi-rigid & 7 & 0.16 & 4 & 0.72 & 6 & 0.25 \\
WEXREY & Semi-rigid & 1 & 0.45 & 1 & 0.19 & 1 & 0.45 \\
\addlinespace[3pt]
Target X & Flexible & NR & -- & 4 & 0.16 & NR & -- \\
Target XIV & Flexible & 1 & 0.38 & NR & -- & 1 & 0.18 \\
Target XVIII & Flexible & NR & -- & NR & -- & NR & -- \\
\bottomrule
\end{tabular}%
\end{table}

\clearpage
\begin{table}[H]
\centering
\small
\caption{\textbf{Complete FastCSP multi-polymorph results.}
Each method reports the energy rank and $\mathrm{RMSD}_{30}$ (\AA) of the
best candidate matching each experimental form. NR denotes no recovery, with
the corresponding $\mathrm{RMSD}_{30}$ unavailable.}
\label{tab:fastcsp-multi-ranking-details}

\footnotesize
\setlength{\tabcolsep}{3pt}
\renewcommand{\arraystretch}{1.05}

\begin{tabular}{@{}lllrrrrrr@{}}
\toprule
& & & \multicolumn{2}{c}{CLARI-H}
& \multicolumn{2}{c}{\model{}-M}
& \multicolumn{2}{c}{\model{}-L} \\
\cmidrule(lr){4-5}\cmidrule(lr){6-7}\cmidrule(l){8-9}
System & Refcode & Form
& Rank & $\mathrm{RMSD}_{30}$
& Rank & $\mathrm{RMSD}_{30}$
& Rank & $\mathrm{RMSD}_{30}$ \\
\midrule
Target I & \texttt{XULDUD01} & Monoclinic
& NR & -- & 1 & 0.297 & 1 & 0.267 \\
Target I & \texttt{XULDUD} & Orthorhombic
& 18 & 0.308 & NR & -- & 18 & 0.270 \\
\addlinespace[3pt]

CILJIQ & \texttt{CILJIQ} & Form I
& 1 & 0.200 & 1 & 0.252 & 1 & 0.256 \\
CILJIQ & \texttt{CILJIQ01} & Form II
& 13 & 0.165 & 11 & 0.118 & 11 & 0.117 \\
\addlinespace[3pt]

Glycine & \texttt{GLYCIN16} & gamma
& 6 & 0.191 & 5 & 0.183 & 5 & 0.187 \\
Glycine & \texttt{GLYCIN20} & alpha
& 1 & 0.128 & 1 & 0.115 & 1 & 0.121 \\
Glycine & \texttt{GLYCIN32} & beta
& 3 & 0.177 & 7 & 0.191 & 4 & 0.194 \\
Glycine & \texttt{GLYCIN67} & delta
& 2 & 0.279 & 2 & 0.280 & 2 & 0.281 \\
Glycine & \texttt{GLYCIN68} & epsilon
& NR & -- & 18 & 0.393 & 14 & 0.379 \\
\addlinespace[3pt]

Imidazole & \texttt{IMAZOL06} & alpha
& 1 & 0.220 & 1 & 0.174 & 1 & 0.284 \\
Imidazole & \texttt{IMAZOL25} & beta
& NR & -- & NR & -- & NR & -- \\
\addlinespace[3pt]

Nicotinamide & \texttt{NICOAM03} & alpha
& NR & -- & 2 & 0.894 & 4 & 0.842 \\
Nicotinamide & \texttt{NICOAM07} & epsilon
& 2 & 0.234 & NR & -- & NR & -- \\
Nicotinamide & \texttt{NICOAM17} & iota
& 10 & 0.781 & 4 & 0.449 & 6 & 0.585 \\
\addlinespace[3pt]

Target VI & \texttt{UJIRIO} & Form I
& 1 & 0.147 & 1 & 0.224 & 1 & 0.185 \\
Target VI & \texttt{UJIRIO05} & Form III
& NR & -- & NR & -- & NR & -- \\
\addlinespace[3pt]

Target XXXI & \texttt{ZEHFUR} & Form B
& NR & -- & 9 & 0.374 & NR & -- \\
Target XXXI & \texttt{ZEHFUR02} & Form A major
& NR & -- & NR & -- & 4 & 0.204 \\
Target XXXI & \texttt{ZEHFUR02} & Form A minor
& NR & -- & NR & -- & 4 & 0.204 \\
\addlinespace[3pt]

ROY & \texttt{QAXMEH01} & Y
& NR & -- & NR & -- & NR & -- \\
ROY & \texttt{QAXMEH12} & YT04
& NR & -- & NR & -- & NR & -- \\
ROY & \texttt{QAXMEH81} & R
& 2 & 0.446 & 3 & 0.473 & 5 & 0.565 \\
ROY & \texttt{QAXMEH03} & OP
& 188 & 0.121 & 195 & 0.247 & NR & -- \\
ROY & \texttt{QAXMEH32} & ON
& 131 & 0.597 & 187 & 0.605 & 153 & 0.570 \\
ROY & \texttt{QAXMEH04} & YN
& 883 & 0.555 & NR & -- & NR & -- \\
ROY & \texttt{QAXMEH05} & ORP
& 1 & 0.354 & 1 & 0.154 & 2 & 0.120 \\
ROY & \texttt{QAXMEH59} & PO13
& 108 & 0.211 & 96 & 0.176 & 106 & 0.165 \\
ROY & \texttt{QAXMEH53} & Y04
& NR & -- & NR & -- & 325 & 0.423 \\
ROY & \texttt{QAXMEH60} & Y19
& 468 & 0.203 & 350 & 0.134 & 479 & 0.190 \\
\bottomrule
\end{tabular}%
\end{table}

\clearpage
\section{Ablation Dataset Preprocessing}
\label{app:dataset-preprocessing}

This section describes the dataset preprocessing used for our ablation studies in \Cref{sec:ablations}. 
Because public split refcodes were not available at the time, we constructed the 
dataset independently. The preprocessing consists of four stages: filtering, structure 
processing, benchmark exclusion, and deduplication. \Cref{tab:dataset-preprocessing-counts}
reports the number of structures remaining after each stage.

\textbf{Filtering.}
Following OXtal~\citep{jin2025oxtal}, we use CSD entries deposited by May 1,
2025 and retain structures with three-dimensional coordinates, an $R$-factor of at most $9\%$, single-crystal X-ray diffraction at ambient pressure, a non-polymeric structure, and a known space group. We
Niggli-reduce each crystal, add missing hydrogens with the CSD Python API~\citep{sykes2024scripting}, and apply the 512-atom limit.

\textbf{Structure processing.}
For entries with incomplete bond labels, we infer missing bond types
and standardize aromatic and delocalized bonds with the CSD Python API~\citep{sykes2024scripting}. 
For each crystal, we store Cartesian coordinates, the unit cell, atomic numbers, bonds and bond types, 
formal charges, stereochemical annotations, and the space-group label. We generate an independent
3D conformer for each molecular component using RDKit ETKDGv3~\citep{rdkit,wang2020improving}.
If conformer generation fails, we retain the crystal and mark the template as unavailable.

\textbf{Benchmark exclusion.}
To prevent leakage, we exclude structures overlapping with the OXtal Rigid and
Flexible benchmarks~\citep{jin2025oxtal}, CSP5--7~\citep{bardwell2011fifth,
reilly2016sixth,hunnisett2024seventhgeneration}, and the CSD Teaching
Subset~\citep{battle2010applications,lo2026clari}. We first remove entries whose
six-letter CSD refcode family matches a benchmark family. We then remove entries
containing a molecular component found in the benchmark set, using the CSD-provided
component SMILES. Following CLARI~\citep{lo2026clari}, we exclude recognized solvents 
and components with fewer than eight heavy atoms from this overlap check.

\textbf{Deduplication.}
We group the remaining entries by six-letter CSD refcode family and cluster
equivalent structures using pymatgen StructureMatcher~\citep{ong2013pymatgen}
with \texttt{ltol}$=0.2$, \texttt{stol}$=0.3$, and \texttt{angle\_tol}$=5^\circ$. 
Within each cluster, we retain the entry with the lowest reported $R$-factor.

\begin{table}[H]
\centering
\caption{\textbf{Ablation dataset preprocessing counts.}
We report the number of structures remaining after each
stage.}
\label{tab:dataset-preprocessing-counts}
\begin{tabular}{@{}lr@{}}
\toprule
Stage & \# Structures \\
\midrule
Raw CSD entries & 1{,}451{,}367 \\
Filtering and structure preprocessing & 833{,}033 \\
Benchmark-family exclusion & 831{,}018 \\
Benchmark-component exclusion & 798{,}459 \\
Deduplication & 766{,}391 \\
\midrule
Training split & 751{,}023 \\
Validation split & 15{,}368 \\
\bottomrule
\end{tabular}
\end{table}
\clearpage
\section{Detailed Architecture Ablation Results}
\label{app:detailed-ablation-tables}

We report the parameter count, GFLOPs, and best validation $\mathrm{Sol}_C$ for
each architecture variant evaluated in \Cref{sec:recipe-architecture}.

\begin{table}[H]
\centering
\caption{\textbf{Architecture ablation with model size and compute.}
We report parameter count and GFLOPs for each architecture variant, together with
its best validation $\mathrm{Sol}_C$ under the shared $\leq300$-atom,
30-candidate protocol with 200-step EDM--Heun sampling. Config.~(F) is the selected cacheable post-entry design.}
\label{tab:architecture_ablation}
\begin{tabular}{lcccccccc}
\toprule
Config. & Entry & TriMul & TriAtt & SU & GEM & $\mathrm{Sol}_C \uparrow$ &
Params (M) & GFLOPs \\
\midrule
(A) & -- & -- & -- & -- & -- & 0.421 & 85.98 & 41.2 \\
(B) & -- & -- & -- & -- & \checkmark & 0.432 & 86.01 & 45.7 \\
(C) & pre & \checkmark & -- & -- & -- & 0.479 & 87.56 & 379.6 \\
(D) & pre & \checkmark & -- & \checkmark & -- & 0.469 & 105.40 & 391.8 \\
(E) & pre & \checkmark & -- & -- & \checkmark & 0.476 & 87.59 & 384.2 \\
\rowcolor{mfTealFill}
(F) & post & \checkmark & -- & -- & -- & 0.474 & 87.56 & 379.6 \\
(G) & post & \checkmark & -- & \checkmark & -- & 0.445 & 105.40 & 391.8 \\
(H) & post & \checkmark & -- & -- & \checkmark & 0.474 & 87.59 & 384.2 \\
(I) & post & \checkmark & \checkmark & \checkmark & -- & 0.462 & 106.07 & 621.1 \\
\bottomrule
\end{tabular}
\end{table}


\end{document}